\documentclass[a4paper,fleqn]{cas-dc}

\usepackage{cite}
\usepackage[authoryear,round]{natbib}
\usepackage{setspace}
\usepackage{subfigure}
\usepackage{pifont}
\usepackage{soul}
\usepackage{lineno}

\let\oldequation\equation
\let\oldendequation\endequation

\renewenvironment{equation}{\linenomathNonumbers\oldequation}{\oldendequation\endlinenomath}

\def\tsc#1{\csdef{#1}{\textsc{\lowercase{#1}}\xspace}}
\tsc{WGM}
\tsc{QE}

\begin{document}
\let\WriteBookmarks\relax
\def\floatpagepagefraction{1}
\def\textpagefraction{.001}

\shorttitle{Efficient Segmentation with Texture in Ore Images Based on Box-supervised Approach}    

\shortauthors{Sun et al.}

\title[mode = title]{Efficient Segmentation with Texture in Ore Images Based on Box-supervised Approach}

\tnotemark[1]


\author[1,2]{Guodong~Sun}
\credit{}

\author[1,2]{Delong~Huang}
\credit{}

\author[1,2]{Yuting~Peng}
\credit{}

\author[1,2]{Le~Cheng}
\credit{}

\author[3]{Bo Wu}
\credit{}

\author[1,2,4]{Yang~Zhang}
\cormark[1] 
\credit{}

\affiliation[1]{organization={School of Mechanical Engineering},
            addressline={Hubei University of Technology}, 
            city={Wuhan},
            postcode={430068}, 
            country={China}}
\affiliation[2]{organization={Hubei Key Laboratory of Modern Manufacturing Quality Engineering},
            addressline={Hubei University of Technology}, 
            city={Wuhan},
            postcode={430068}, 
            country={China}}
\affiliation[3]{organization={Shanghai Advanced Research Institute},
			addressline={Chinese Academy of Sciences}, 
			city={Shanghai},
			postcode={201210}, 
			country={China}}
\affiliation[4]{organization={National Key Laboratory for Novel Software Technology},
			addressline={Nanjing University}, 
			city={Nanjing},
			postcode={210023}, 
			country={China}}
\cortext[1]{Corresponding author: Yang Zhang} 
\tnotetext[1]{yzhangcst@hbut.edu.cn (Yang Zhang)}

\begin{abstract}
Image segmentation methods have been utilized to determine the particle size distribution of crushed ores. Due to the complex working environment, high-powered computing equipment is difficult to deploy. At the same time, the ore distribution is stacked, and it is difficult to identify the complete features. To address this issue, an effective box-supervised technique with texture features is provided for ore image segmentation that can identify complete and independent ores.
Firstly, a ghost feature pyramid network (Ghost-FPN) is proposed to process the features obtained from the backbone to reduce redundant semantic information and computation generated by complex networks. Then, an optimized detection head is proposed to obtain the feature to maintain accuracy. Finally, Lab color space (Lab) and local binary patterns (LBP) texture features are combined to form a fusion feature similarity-based loss function to improve accuracy while incurring no loss. Experiments on MS COCO have shown that the proposed fusion features are also worth studying on other types of datasets. Extensive experimental results demonstrate the effectiveness of the proposed method, which achieves over 50 frames per second with a small model size of 21.6 MB. Meanwhile, the method maintains a high level of accuracy 67.8 in~$AP_{50}^{box}$ and 47.7 in~$AP_{50}^{mask}$ compared with the state-of-the-art approaches on ore image dataset, even better than bounding box tightness prior (BBTP) by 10.4/1.3 on ~$AP_{50}^{box}$/~$AP_{50}^{mask}$metrics with the ResNet50 as backbone. The source code is available at \url{https://github.com/MVME-HBUT/OREINST}.
\end{abstract}




\maketitle


\section{Introduction}
Using artificial intelligence technology, smart mining aspires to create a digital and information-based industry, with ore image processing fulfilling many complicated and risky activities automatically and effectively~\citep{EAAI6}. Ore particle size detection is a crucial task in the mining process. The anomalous particle size distribution represents mining equipment failure and potential safety issues in the mining production process. Accurate image segmentation is the foundation for obtaining physical information about an object \citep{eaai1, eaai2, eaai3}, and ore is no exception. Since the detection equipment is typically arranged outside, light and dust always have an impact on the quality of the captured images. The accuracy of ore image segmentation is easily affected by the characteristics of the ore itself, the texture, and the stacking of ores. Due to environmental limitations in the field, only constrained resource equipment is accessible for practical application. 

\begin{figure*}[!t]
	\centering
	
	\subfigure[]{
		\begin{minipage}[c]{0.18\linewidth}
			\includegraphics[width=1.1in]{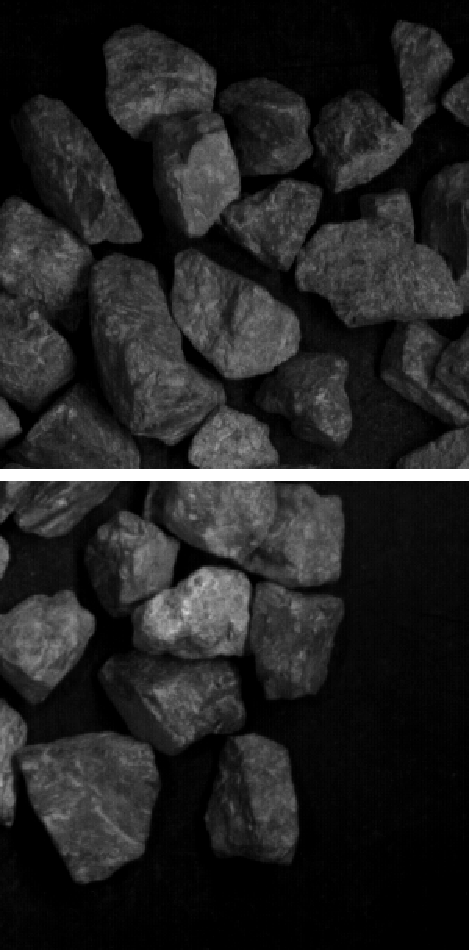}\vspace{0.4em} 
		\end{minipage}
	}\hspace{-1em}
	\subfigure[]{
		\begin{minipage}[c]{0.18\linewidth}
			
			\includegraphics[width=1.1in]{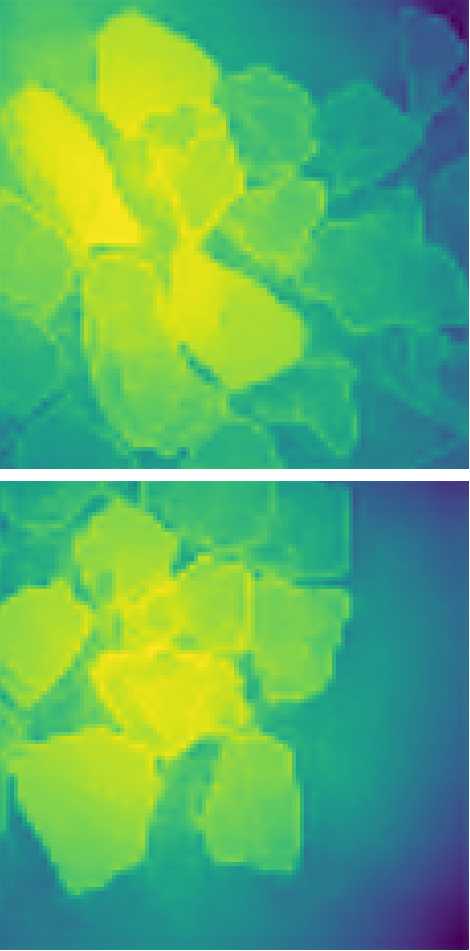}\vspace{0.4em} 

		\end{minipage}
	}\hspace{-1em}
	\subfigure[]{
		\begin{minipage}[c]{0.18\linewidth}
			
			\includegraphics[width=1.1in]{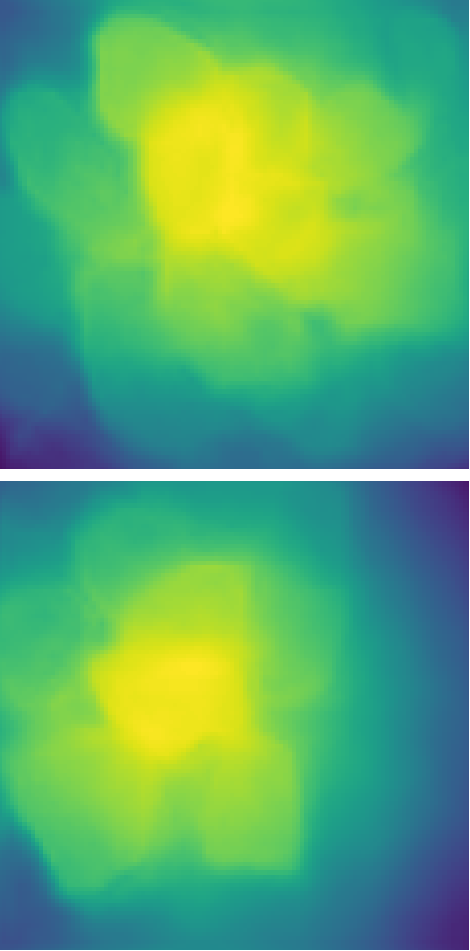}\vspace{0.4em} 

		\end{minipage}
	}\hspace{-1em}
	\subfigure[]{
		\begin{minipage}[c]{0.18\linewidth}
			
			\includegraphics[width=1.1in]{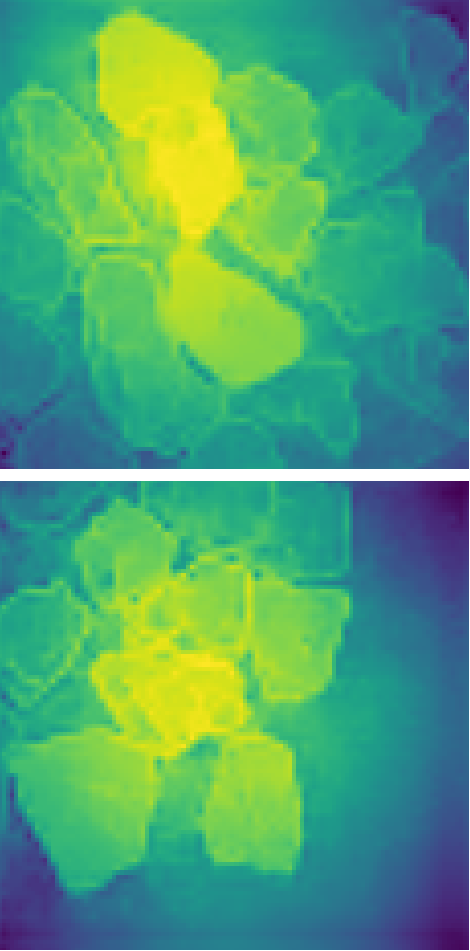}\vspace{0.4em} 

		\end{minipage}
	}
	\caption{The attention maps of the network based on different features. High color brightness indicates a high level of attention. (a) Input images. (b), (c) and (d) The focus positions of Lab color space (Lab), local binary patterns (LBP) texture feature, and the proposed fusion feature on ore images, respectively. When there is only Lab, the attention of the network is easily spread over numerous ores. Due to the restricted color distribution range of ore images, it causes a huge mistake in feature similarity computation. The concerned position of the network is led to the proper location with the assistance of the LBP texture feature.
	}
	\label{intro1}
\end{figure*}

To accomplish the task of ore image segmentation, researchers initially attempt to apply traditional image processing methods such as the watershed algorithm~\citep{2011Automatic}, the fogbank algorithm~\citep{2014FogBank}, and a regression-based classifier~\citep{2009Ore}. These algorithms can segment ore images to a certain extent, but it is challenging to obtain higher accuracy under complex conditions. Moreover, the robustness and complexity of these traditional methods are lacking.  Convolutional neural network (CNN) based algorithms have been applied to practical engineering since the emergence of deep learning technology \citep{EAAI5,EAAI4,EAAI6,ann}. Wang et al.~(\citeyear{WangWei}) proposed an ore image segmentation approach based on U-Net with a boundary mask fusion block. However, the computational cost of these complex CNN-based methods is extremely high, which typically identifies ore edge rather than separating each independent ore. 

In CNN-based methods, instance segmentation has a broad application prospect and is more aligned with ore segmentation needs. In industrial quality inspection, the two-stage method Mask-RCNN \citep{Mask-RCNN} and the one-stage method CondInst~\citep{CondInst} produced good results. In practical work environments, the production of dataset is extremely challenging when using these fully-supervised instance segmentation methods. Tian et al.~(\citeyear{BoxInst}) proposed a box-supervised instance segmentation method based on CondInst, which utilizes the Lab color space (Lab) to calculate the similarity between pixel pairs. For ore images, this method disregards the texture feature of objects and still has flaws such as being too slow or having a large model size. To achieve a balance of speed, accuracy, and model size in a resource-constrained environment, a lightweight and high-precision framework is required to meet the actual needs of real-time speed and small model size.

To address these problems, a lightweight and accurate framework is presented for performing instance segmentation of ore images. The proposed network can be trained using the box-only supervision method. A fusion feature that combines color space and texture features is proposed, which can take into account both local and global information about the ore. By analyzing the feature, the detection accuracy can be greatly improved.
First, the lightweight backbone MobileNetv3-small is used to reduce computation cost and model size, and a ghost feature pyramid network (Ghost-FPN) is proposed to reduce computation while keeping sufficient semantic information. Then, an optimized detection head is proposed to eliminate extraneous semantic information and computational expense. 
Finally, a loss function based on fusion features is proposed, which can self-supervised the training mask without relying on mask annotations. Fig.~\ref{intro1} depicts attention maps with various features. The highlighted parts in the fusion feature attention map are more concentrated and accurate when the Lab color space (Lab) and local binary patterns (LBP) texture features are compared. Experimental results on the MS COCO dataset demonstrate the importance and effectiveness of the LBP texture feature. In the ablation experiment, the fusion features of different weight ratios were compared, and the highest precision weight ratio was selected for model training. On ore image dataset, experiments reveal that the proposed network outperforms the state-of-the-art methods with a model size (21.6 MB) and real-time speed (19.8 ms).
In general, the main contributions are as follows.
\begin{enumerate}[1)]
	\item For loss function, a novel fusion feature similarity-based is designed, which includes Lab color space (Lab) mixed with local binary patterns (LBP) texture feature to improve accuracy without negatively affecting inference speed and model size.
	\item For lightweight and real-time, a Ghost-FPN and an optimized detection head are presented in the proposed box-supervised approach.
	\item Experiments on the MS COCO and ore image datasets show that the proposed fusion feature enhances the network, offering a workable improvement strategy for networks involving color space and texture features.
\end{enumerate}  

The rest of this article is organized as follows. Section~\ref{sec:related_works} introduces related works about ore image segmentation methods, fully-supervised, and box-supervised instance segmentation methods. Section{~\ref{sec:method}} presents the framework with two individual modules and the novel loss function. To verify the effectiveness of the proposed approach, comprehensive experiments are shown in Section{~\ref{sec:experiments}}. Finally, the full text is summarized in Section~\ref{sec:conclusion}.

\section{Related works}\label{sec:related_works}
\subsection{Ore Image Segmentation}
Researchers have investigated numerous ore image segmentation approaches, ranging from traditional methods to machine learning and deep learning. 
For example, Mukherjee et al. (\citeyear{2009Ore}) utilized a regression-based classifier to learn ore shape features, which improved the segmentation accuracy of ore boundary, but the parameters were manually adjusted.
In traditional image segmentation algorithms, watershed~\citep{2011Automatic} was often used in region-based segmentation techniques. However, it was difficult to accurately segment the ore particles with fuzzy edges, uneven - illumination, and adhesion degree.
Wang et al. (\citeyear{WangWei}) proposed an ore image segmentation approach based on deep learning-based methods. However, the computing cost of these methods was prohibitively expensive. Although the segmentation method of ore images was improved, there are still some major issues, such as low precision, slow speed, large model size, and strict constraints for use, which the aforementioned methods did not address.
\subsection{Fully-supervised Instance Segmentation}
Existing approaches are classified into two types: two-stage and one-stage. The detect-then-segment paradigm is followed by two-stage approaches. Mask R-CNN~\citep{Mask-RCNN} extended Faster R-CNN~\citep{Faster-RCNN} by including a full convolution network (FCN) mask branch. Mask scoring R-CNN~\citep{Mask-Scoring-RCNN} based on Mask R-CNN addressed the misalignment issue between mask quality and classification score. 
Chen et al.{~\citeyearpar{HTC}} proposed a hybrid task cascade (HTC) that connected the box and mask branches in a multi-stage cascaded fashion, allowing the network to better distinguish objects from disorderly backgrounds. CARAFE \citep{CARAFE} enabled instance-specific content-aware handling, which generated adaptive kernels on-the-fly, rather than using a fixed kernel for all samples. The performance of the network is improved by increasing the amount of computation and parameters by a small amount. One-stage techniques include mask prediction into a simple FCN-like architecture without RoI cropping. Bolya et al.~\citeyearpar{YOLACT} proposed a method named you only look at coefficients (YOLACT), which divided the instance into two simultaneous jobs, created a set of prototype masks, and predicted the masking coefficient for each instance. At the same time, the article claims that Fast-NMS greatly improves segmentation speed. 
BlendMask {\citep{BlendMask}} combined top-down and bottom-up approaches, outperforming Mask R-CNN under the same training schedule while being 20\%. 
SOLOv2 {\citep{SOLOv2}} generated full instance masks by location without detection, which uses a novel matrix non-maximum suppression (NMS) technique to reduce inference overhead significantly. CondInst {\citep{CondInst}}, unlike Mask R-CNN, did not rely on ROI operations but instead employed conditional convolutions to predict instance-aware masks, making it faster. These methods outperform in terms of accuracy, but they still suffer from large model sizes and slow speeds for large datasets. In practice, these methods require a significant amount of time to generate dataset.

\subsection{Box-supervised Instance Segmentation}
In deep learning, semi-supervised{~\citep{mglnn}} has also developed rapidly, and weakly supervised instance segmentation with box annotations has received little attention. Simple Does It {\citep{SDI}} was the first instance of segmentation with box annotations, which refined the segmentation results using an iterative training program based on the region suggestions given by multiscale combinatorial grouping. These approaches treated each box with annotation separately, with no uniformity in the annotation. Mask R-CNN was the foundation of bounding box tightness prior (BBTP)~{\citep{BBTP}}. The multi-instance learning task was used to sample positive and negative bags according to RoI on the CNN-based feature map, thus solving the problem of unpredictable mask distribution caused by unsupervised signals in the box, whose performance is ahead of Simple Does It. However, the important previous information received from the color of the near-point pixel was not employed in BBTP. BoxInst \citep{BoxInst} was based on CondInst without RoI, and the proposed projection loss term was utilized to monitor mask learning, eliminating the requirement for sampling. BoxInst improved the accuracy and speed of box-supervised methods significantly. When more semantic correspondence was desired, DiscoBox \citep{discobox} employed bounding box supervision to jointly learn instance segmentation and semantic correspondence, and the result outperformed existing weak supervision methods. However, it necessitates more training memory, training time, and a larger model size. These methods address the difficulty in dataset production, but they still face significant challenges in actual work environments.

\begin{figure*}[!t]
	\centering
	\includegraphics[width=6.8in]{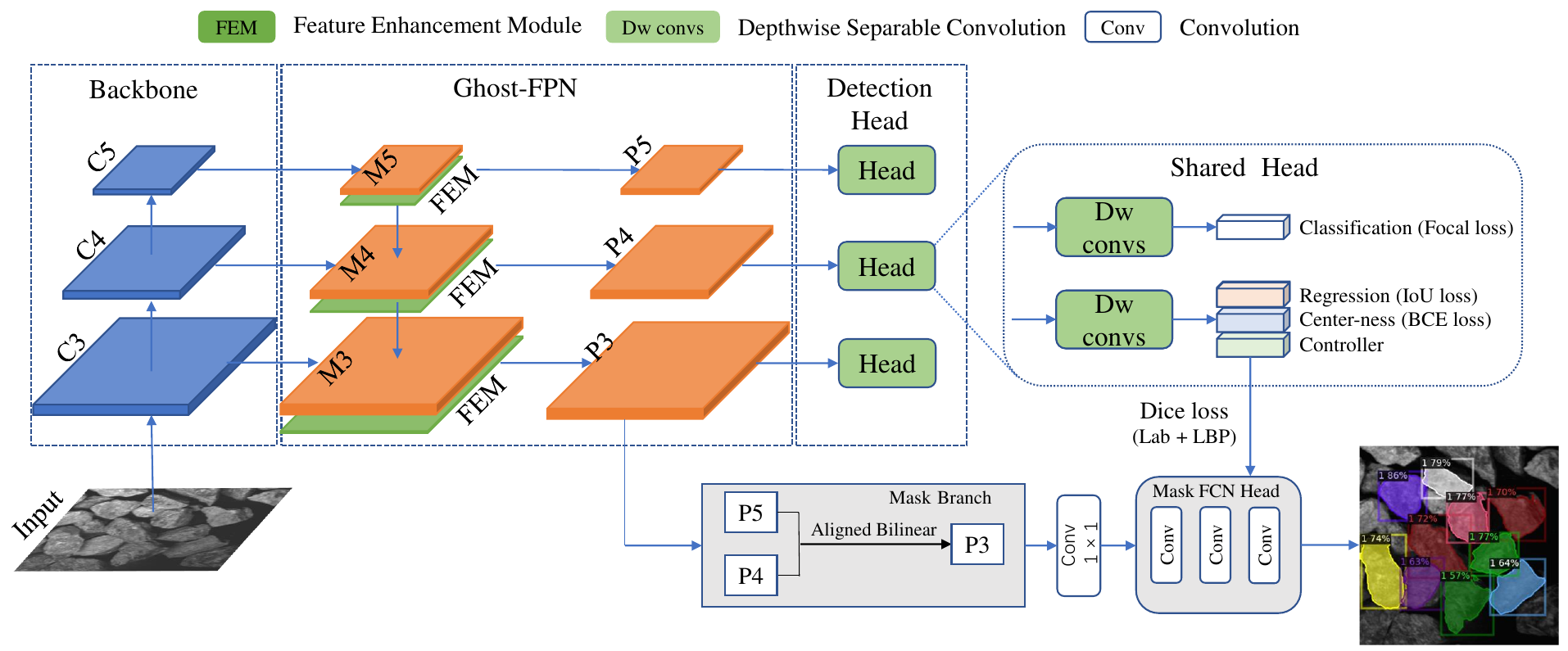}
	\caption{Architecture of the proposed OreInst. 
		The proposed framework includes three parts: the backbone, the ghost feature pyramid network (Ghost-FPN), and the detection head. The M3, M4, and M5 are the results for channel alignment of output features in the backbone. The Ghost-FPN and detection head have just 96 channels. The 3$\times$3 convolutional layers are removed from the original FPN, and the ghost block is added as a feature enhancement module (FEM) to create a novel Ghost-FPN. The three scale feature maps generated by ghost-FPN are fed into the detection head for classification and regression tasks. The detection head comprises four branches, and the ordinary convolution in each branch is replaced by depthwise separable convolution. The classified branch performs the classification task, and the location task is performed by the regression branch. The center-ness branch is introduced to suppress the low-quality detected bounding boxes. The controller branch generates the parameters of the mask full convolution network (FCN) head. The number of channels in Ghost-FPN and detection head is 96. In the mask branch, the feature maps P4 and P5 are added to P3 after the linear calculation. These feature maps are input into the mask FCN head to obtain the final result. The mask loss uses dice loss, and pairwise loss employs the superposition of Lab and LBP texture features to calculate the feature similarity.}
	\label{approach}
\end{figure*}

\begin{figure}[!t]
	\centering
	\subfigure[]{
		\includegraphics[width=2.1in]{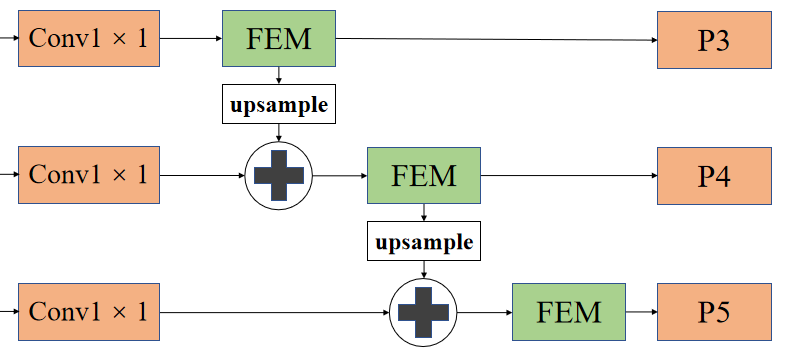}
	}
	\hspace{-0.9em}
	\subfigure[]{
		\includegraphics[width=0.7in]{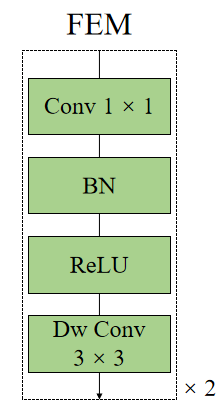}
	}
	\caption{(a) Architecture of Ghost-FPN. 
		A ghost block is added as an FEM to the feature pyramid network (FPN) to create a Ghost-FPN. In Ghost-FPN, the input channel is limited to 96. (b) The architecture of FEM. FEM consists of two identical parts. Each unit includes a 1 $\times$ 1 Convolution, a batch normalization (BN) layer, a ReLU activation function, and a 3 $\times$ 3 depthwise separable convolution.}
	\label{ghostFPN}
\end{figure}

\section{Method}\label{sec:method}
In this section, the overall framework is first introduced. Following that, the proposed Ghost-FPN and optimized detection head are presented. Finally, the loss function in the framework is described. Simultaneously, a novel fusion feature is designed that combines color space and texture features.

\subsection{Overall Framework}
As discussed before, hardware resources are limited in the field. Nowadays, the deep learning research method based on Transformer is widely used in the industrial field {\citep{pcgvit,densesph-yolov5}}. Although the method based on the transformer has high precision, it takes up a lot of memory, so the CNN-based method is more suitable for the application scenario of this article. If the CNN-based approaches are applied to segment the ore images in the field accurately, the instance segmentation network must have a low computational cost. A large portion of the computation is often performed in the backbone, neck, and detection head. These three parts are optimized for ore images so that the network retains competitive accuracy while using the least amount of computational cost. A real-time instance segmentation approach is proposed for ore images (OreInst) that consists primarily of a lightweight backbone, a Ghost-FPN, and an optimized detection head.

To reduce computation, MobileNetv3-small~\citep{MobileNetV3-Small} is used as the backbone.  In Fig.~\ref{approach}, the backbone generates feature maps across different scales. High-level feature maps provide more semantic information, while low-level have more resolution. The Ghost-FPN is programmed with features in different scales. Rich semantic information of upper-layer feature maps is communicated to the lower-layer for fusion. Before fusion, a feature enhancement module (FEM) treats the feature map from each layer. The FEM improves the fusion of semantic information at various scales. Simultaneously, the FEM employs depthwise separable convolutions internally, adding a minor computational burden to the network to compensate for the semantic information loss caused by the reduced number of channels. The fusion feature maps are fed into the detection head, which connects the following output heads. The classification head is used to predict the category of a certain location. The regression head is responsible for locating and regressing the prediction target. Especially the center-ness head is used to suppress the discovered low-quality bounding boxes. The controller's head anticipates mask head parameters for the instance at the location. The four branches in the detection head comprise two 5 $\times$ 5 depthwise separable convolutions (exclude the final prediction layers). Compared with other networks, the convolutional layer in the detection head has a larger receptive field, allowing it to perform better in classification and regression tasks. Simultaneously using deep depthwise convolution keeps the computational and parameter quantities at a low level. To acquire the results, P3, P4, and P5 layers are translated to mask branches, which are convoluted and passed to the mask FCN head. During the network training process, fusion features are applied in the loss function, making the network more accurate. Based on the design of various network components described above, the final network maintains a high level of accuracy while outperforming existing networks in terms of inference speed, model size, and memory usage.

Block-combined neural networks and block-based neural networks (BbNNs), which were developed by numerous researchers, have excelled in their respective subdivision tasks \citep{Block_network1,Block_network2,Block_network3}. And when transferred to other networks, the designed module might still produce good results. Similar results are obtained when other neural networks are combined with the Ghost-FPN, which was created as a neck. However, this still requires scientific and reasonable experiments to prove. Efficient block-wise neural network architecture Generation proposed by Zhong et al. \citeyearpar{Block_network4} can create neural networks made up of numerous blocks. However, the generated networks perform worse than the networks designed by the researchers in the majority of tasks.

\subsection{Ghost-FPN}
A novel Ghost-FPN is suggested to reduce the model size and improve inference speed while maintaining accuracy. The feature maps with strong semantic information are added to the relevant feature maps in the original FPN. To obtain the P3, P4, and P5 layers, these feature maps are subjected to a 3$\times$3 convolutional layer. From P5, two downsamples are obtained to produce P6 and P7, respectively. It is commonly established that features of varying scales have a greater influence on spotting objects of varying sizes. The ores have been treated using an ore crusher, showing that the size of these ores is manageable. In this case, small-scale feature maps are no longer required. So the layers P6 and P7 from FPN are eliminated. The following experiments show that the appeal operation has a positive effect on the network.

After removing the network module that creates redundant information, the number of calculations and parameters in the FPN still needs to be reduced while keeping the accuracy loss within an acceptable range. Based on the above requirements, a Ghost-FPN is proposed as shown in Fig. \ref{ghostFPN}(a). The bigger the number of convolution channels for ore images, the higher the accuracy of the network will not continue to rise. As a result, the optimum number of channels is critical to the lightweight of the network. Because the number of channels must be kept at a multiple of 16, the framework can benefit from the maximum parallel acceleration. The 96 channels are a trade-off between accuracy and model size. Then, in each layer of FPN, the standard 3$\times$3 convolutions of output features are removed, reducing computation and parameters. Ghost block \citep{GhostNet} is presented as FEM to compensate for the loss of regular convolution, as shown in Fig. {\ref{ghostFPN}(b)}. The basic structural unit of a ghost block comprises 1$\times$1 standard convolution and 3$\times$3 depthwise separable convolution, and each ghost block contains two primary units. Considering the following changes in convolutional layer number, channel number, and convolution kernel size, the computation of standard convolution and depthwise separable convolution is calculated as follows:
\begin{equation}\label{FPNequ1}
\begin{aligned}
	&S_{conv} = K \times K \times C_{in} \times W_{out} \times H_{out} \times C_{out}, \\
	&DW_{conv} = K \times K \times C_{in} \times W_{out} \times H_{out} \\
	&+ C_{in} \times W_{out} \times H_{out} \times C_{out},\\
\end{aligned}
\end{equation}
where $ K $, $ C_{in} $, and $ C_{out} $ denote the kernel size, number of input channels, and number of output channels, respectively. $ W_{out} $ and $ H_{out} $ denote the width and height of the output feature map, respectively. Because $ W_{out} $ and $ H_{out} $ are the same, the number of calculations and parameters of the optimized part in FPN is $ 1/15 $ of the original.

\subsection{Optimized Detection Head}

\begin{figure}[!t]
	\centering
	\subfigure[]{              
		\includegraphics[width=1.5in]{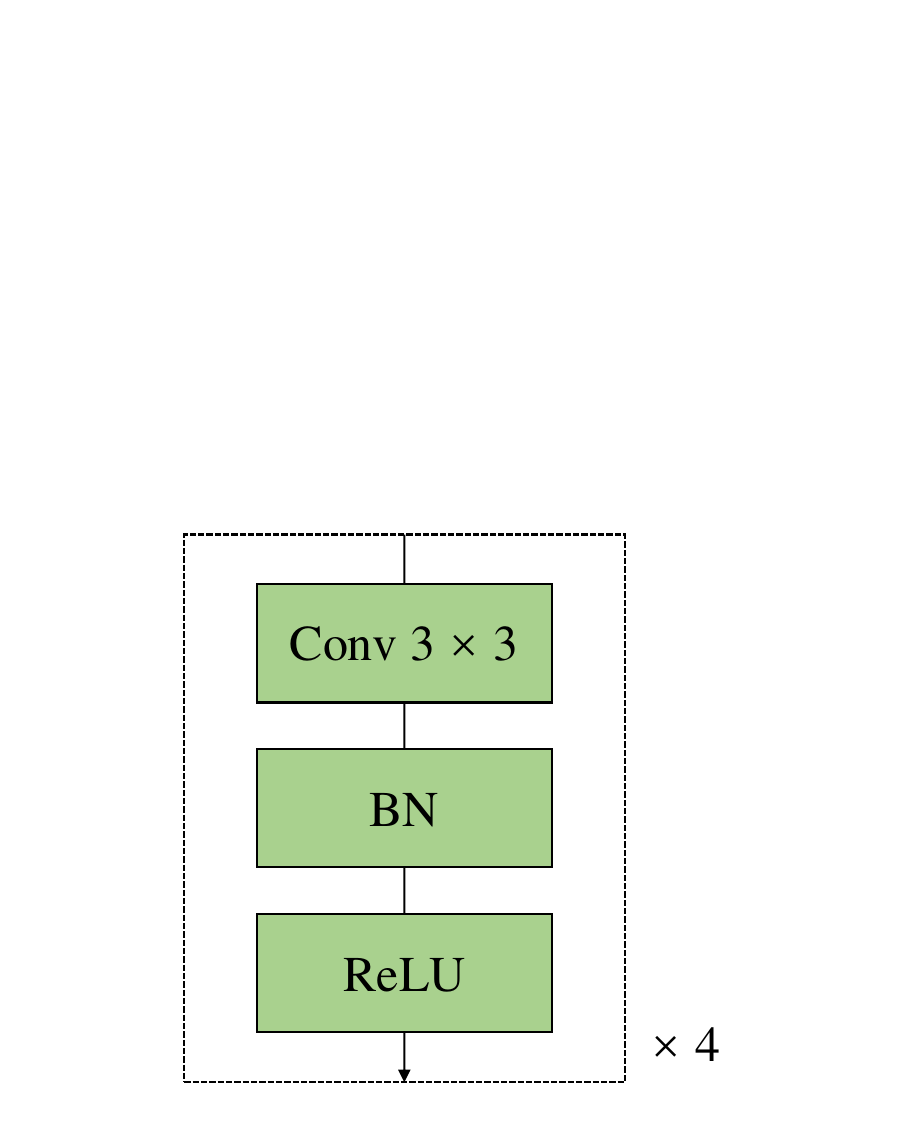}
	}
	\hspace{-1em}
	\subfigure[]{
		\includegraphics[width=1.5in]{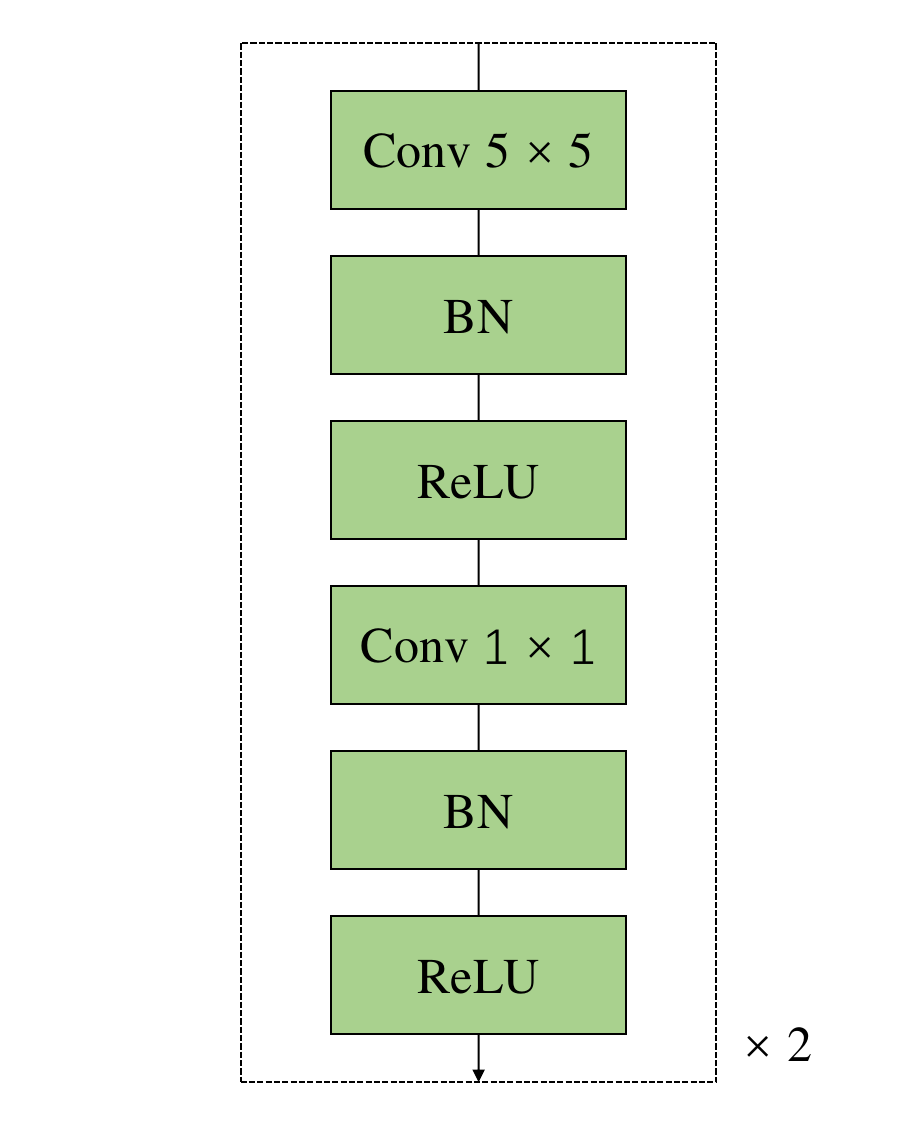}
	}
	\caption{(a) Original detection Head and (b) the optimized detection head.}
	\label{DWconv}
\end{figure}

\begin{figure}[!t]
	\centering
	\subfigure[]{
		\begin{minipage}[b]{0.3\linewidth}
			\centering
			\includegraphics[width=1.0in]{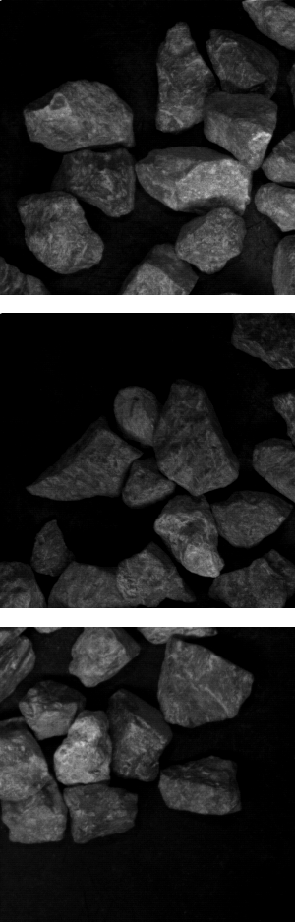}\vspace{0.5em} 

		\end{minipage}
	}\hspace{-0.6em}
	\subfigure[]{
		\begin{minipage}[b]{0.3\linewidth}
			\centering
			\includegraphics[width=1.0in]{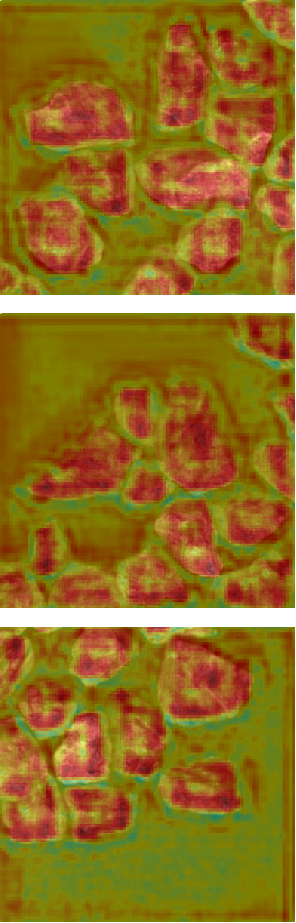}\vspace{0.5em}

		\end{minipage}
	}\hspace{-0.6em}
	\subfigure[]{
		\begin{minipage}[b]{0.3\linewidth}
			\centering
			\includegraphics[width=1.0in]{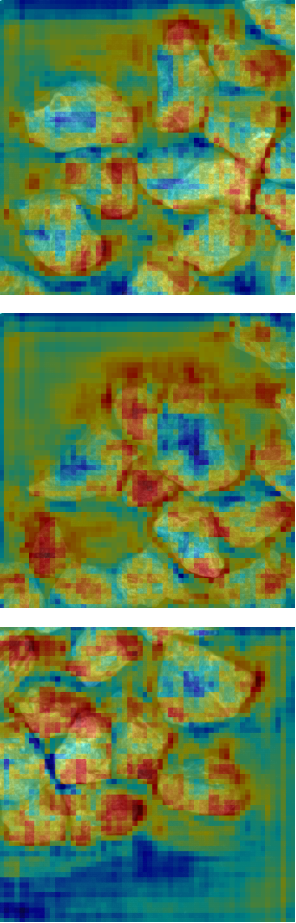}\vspace{0.5em} 

		\end{minipage}
	}
	\caption{The color close to red in the figure denotes strong attention, whereas the color close to blue denotes low attention. (a) Input images. (b) The feature maps output from the original framework. (c) The feature maps output from the proposed framework.}
	\label{headfeaturemap}
\end{figure}

Since the input and output channels are adjusted of FPN to 96, the input channel of the detection head is also naturally changed to 96. To further reduce computation, the network structure of the detection head is simplified. First, two groups of shared convolution are used instead of four groups of convolution in two branches of box regression and classification.
To run on hardware with limited resources, using shared convolution reduces the computation amount and improves the running speed. In addition, other types of convolution are chosen to replace traditional convolution. In the same input case, the output of a depthwise separable and traditional convolution is the same. But the former is more efficient than the latter. As shown in Fig. \ref{DWconv}(a), the traditional convolution consists of a 3$\times$3 convolution, a batch normalization (BN) layer, and a rectified linear unit (ReLU) layer. Depthwise separable convolution as shown in Fig. \ref{DWconv}(b) adds a 1$\times$1 convolution, BN, and ReLU layers. 

To improve the receptive field and performance of the detector, the 3$\times$3 convolution in depthwise separable convolution is then replaced by a 5$\times$5 convolution, which requires only a small amount of computation. Nevertheless, the number of parameters and multiplications of the optimized detection head is far less than that of the original. 
According to Eq.~(\ref{FPNequ1}), because $ W_{out} $ and $ H_{out} $ are the same, the number of calculations and parameters of the optimized part in detection head is $ 0.5\% $ of the original.
Figure \ref{headfeaturemap} (b) shows that the original framework cannot discriminate between ore and background. In Fig.~\ref{headfeaturemap} (c),  the proposed framework makes it easier to obtain results close to the ground-truth.

\subsection{Loss Function}
The proposed loss function can be divided into a projection loss term and a pairwise affinity loss term. These two loss functions can self-supervise the training mask without relying on the mask annotation.

\subsubsection{Projection Loss Term} 
The first term uses the ground-truth box annotation to supervise the horizontal and vertical projections of the predicted mask, ensuring that the tightest box encompassing the predicted mask matches the ground-truth box. The following is the definition of the projection loss term:
\begin{equation}\label{loss_1}
	L_{proj} = L(\tilde{\textbf{l}}_x,\textbf{l}_x) + L(\tilde{\textbf{l}}_y,\textbf{l}_y),
\end{equation}
where $ \textbf{l}_x $ and $ \textbf{l}_y $ indicate that the mask is on the x-axis and y-axis, respectively. The \emph{L} is the dice loss as in CondInst~\citep{CondInst}. For the predicted mask, the corresponding projections {$ \tilde{\textbf{l}}_x $} and {$\tilde{\textbf{l}}_y $} could also be obtained.

\begin{figure}[t]
	\centering
	\subfigure[]{
		\includegraphics[width=1.2in]{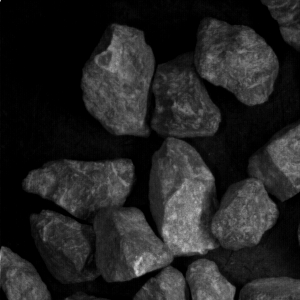}
	}\hspace{-0.6em}
	\subfigure[]{
		\includegraphics[width=1.2in]{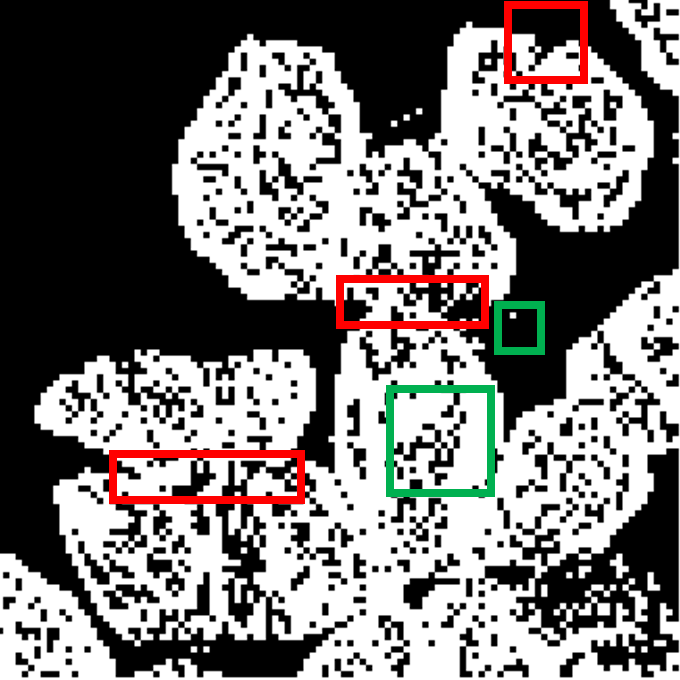}
	}\hspace{-0.6em}	
	\subfigure[]{
		\includegraphics[width=1.2in]{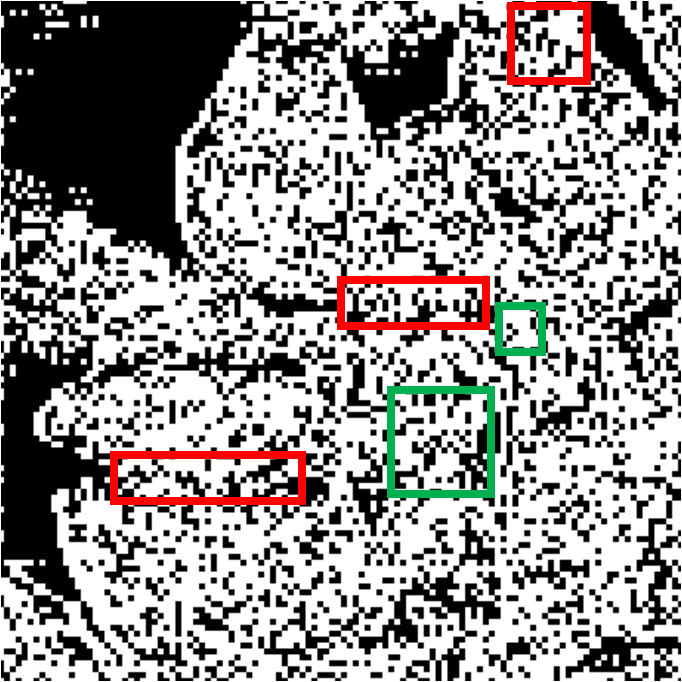}
	}\hspace{-0.6em}
	\subfigure[]{
		\includegraphics[width=1.2in]{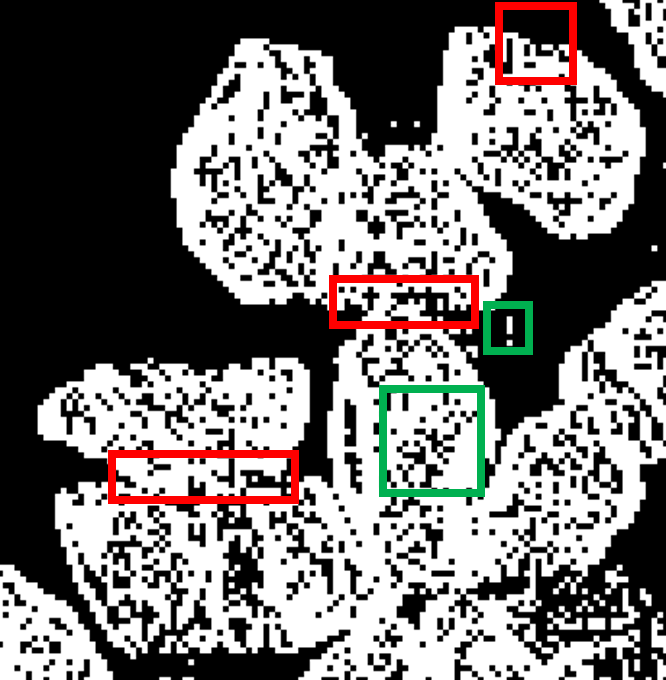}
	}
	
	\caption{The red box and green box represent the positive and negative auxiliary effects of LBP texture feature, respectively.
		(a) Input images. (b), (c) and (d) represent the similarity between pixels on ore images using Lab, LBP texture feature, and the proposed fusion feature, respectively.}
	\label{colorsimilarity}
\end{figure}

\subsubsection{Pairwise Affinity Loss Term}
Recently, almost instance segmentation methods supervised the predicted masks on a per-pixel basis. The pixel-wise supervision is unavailable without the mask annotations. In this case, the mask is supervised in pairs.

Based on the ground-truth masks, an undirected graph \emph{G} = (\emph{P}, \emph{E}) is considered constructed on an image, where \emph{P} is the set of pixels and \emph{E} is the set of edges. 
Each pixel is linked to its \emph{K} $ \times $ \emph{K} $ - $ 1 neighbors by using dilated convolution. 
The label for the edge $ y_e \in {0, 1}$,
where $ y_e = 1 $ indicates that the two pixels connected by the edge have the same ground-truth label, while $ y_e = 0 $ indicates that their labels are different. Let pixels (\emph{i}, \emph{j}) and (\emph{l}, \emph{k}) represent the two endpoints of the edge. The network prediction $ \tilde{\textbf{m}}_{i,j} $ is through as the probability of pixel (\emph{i}, \emph{j})
being foreground. So the probability of $ y_e = 1 $ is:
\begin{equation}\label{loss_2}
	P(y_e = 1) = \tilde{\textbf{m}}_{i,j}\cdot\tilde{\textbf{m}}_{k,l} + (1 - \tilde{\textbf{m}}_{i,j})\cdot(1 - \tilde{\textbf{m}}_{k,l}),
\end{equation}
and $ P(y_e = 1) $ + $ P(y_e = 0) $ = 1. The binary cross entropy loss is commonly used to train the probability distribution resulting from network prediction.
Then, the feature similarity between a pair of pixels is calculated with a feature similarity threshold of $ \tau $. The pixel pair belongs to the same instance if the feature similarity of the pixel pair is above $ \tau $.
Formally, the feature similarity is defined as:
\begin{equation}\label{loss_3}
S_e = S(\textbf{c}_{i,j},\textbf{c}_{l,k}) = exp(-\frac{||\textbf{c}_{i,j} - \textbf{c}_{l,k}||}{2}),
\end{equation}
where $ S_e $ be the feature similarity of the edge, and $ \textbf{c}_{i,j} $ and $ \textbf{c}_{l,k} $ denote the feature vectors of the two pixels (\emph{i}, \emph{j}) and (\emph{l}, \emph{k}) connected by the edge, respectively.
For these confident edges exclusively, the pairwise loss and the labels that {$ S_e $} less than {$ \tau $} can be computed and eliminated. The pairwise loss becomes:
\begin{equation}\label{loss_4}
	L_{pair} = -\frac{1}{N}\sum\limits_{e \in E_{in}}\mathbb I_{{S_e\geqslant\tau}}logP(y_e = 1),
\end{equation}
where $ E_{in} $ is the set of the edges in the box that contains at least one pixel. 
$ \mathbb I_{{S_e\geqslant\tau}} $ is 1 if $ S_e $ $ \geqslant $ $ \tau $, otherwise, it is 0. 
Using $ E_{in} $ instead of \emph{E} means keeping the loss from being dominated by a significant number of pixels outside the box.
\emph{N} is the number of the edges in $ E_{in} $. Overall, the total loss for mask learning can be formulated as $L_{mask} = L_{proj} + L_{pair}$.

\subsubsection{Fusion Feature Similarity} 
Based on Eq. (\ref{loss_3}) and Eq. (\ref{loss_4}), the more accurate the similarity judgment of pixel pairs, the higher the accuracy of the network. Other features of ores are omitted if only the color space is utilized to determine feature similarity. To improve the accuracy of feature similarity, a fusion feature is proposed, integrating color space and texture features.

\textbf{Lab:} The color gamut of Lab color space is extremely broad, even beyond that of a computer screen and human vision. The lab can clearly indicate how each color is formed and presented once the white point of the color space is given, which has nothing to do with the display medium employed. So, the similarity between pixels can be calculated precisely by using only the Lab color space. But because of the unique texture features of the ore surface, the LBP texture feature is introduced as follows.

\textbf{LBP texture feature:} LBP texture feature can efficiently describe the texture features of objects, measure and extract image local texture information, and it is invariant to illumination. The uniform mode in the LBP texture feature has both rotation invariance and illumination invariance. Regarding local texture description, it outperforms the traditional LBP texture feature, which has been successfully applied to face detection, lip recognition, expression detection, dynamic texture, and other fields. The LBP texture feature can calculate the feature similarity of an ore edge accurately.
As shown in Fig. \ref{colorsimilarity}, Lab calculates the feature similarity of most parts of the ore images accurately, but the feature similarity of the ore to background and ore to its boundary is not precise enough. For this reason, the fusion feature is proposed with the LBP texture feature as the auxiliary element and Lab as the main element. The goal is to improve the accuracy of pixel similarity in the above two cases.

In general, the new similarity is calculated as follows:
\begin{equation}\label{loss_6}
	{ S_e}^{\ast} = {\theta}_{1} \cdot S_{e(Lab)} + {\theta}_{2} \cdot S_{e(LBP)},
\end{equation}
where $ S_{e(Lab)} $  represents color similarity estimated using Lab, while $ S_{e(LBP)} $ represents texture similarity derived using LBP texture feature. $ { S_e}^{\ast}  $ denotes feature similarity composed of color and texture similarity under different weights.
The weights of color and texture similarity are $ {\theta}_{1} $ and $ {\theta}_{2} $, respectively. In this work, the values for $ {\theta}_{1} $ and $ {\theta}_{2} $ are 0.9 and 0.1, respectively. When the LBP texture feature is used to calculate the pixel pair similarity of an ore image, the boundary similarity of the ore is typically close to 1. When using Lab to calculate the similarity of the same ore boundary pixel pair, the $ { S_e}^{\ast}  $ will be greater than 0.2 ($ \tau $ = 0.2) as long as the similarity exceeds 1/9. In this case, the similarity of pixel pairs has increased from less than 0.2 to greater than 0.2.
In Fig. \ref{colorsimilarity}, red box appears more on the ore boundary, but green box appears more in the background and the ore. The red box illustrates that the LBP texture feature assists the network in better determining the boundary of minerals when computing the similarity among pixels.

\begin{figure}[!t]
    \centering
    \subfigure[]{
    \includegraphics[width=2.15in]{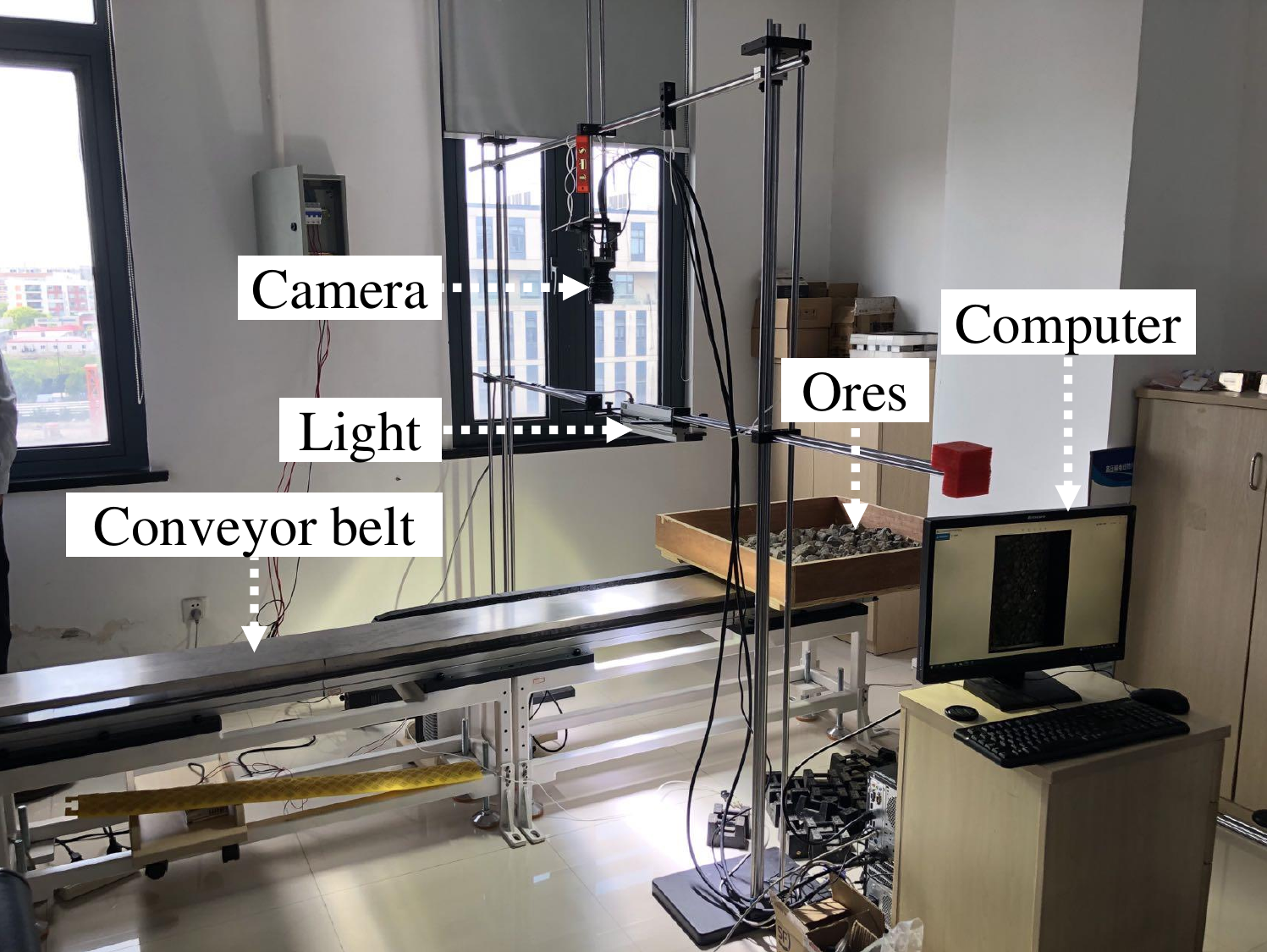}
    \label{platform}}
    \subfigure[]{
    \includegraphics[width=0.78in]{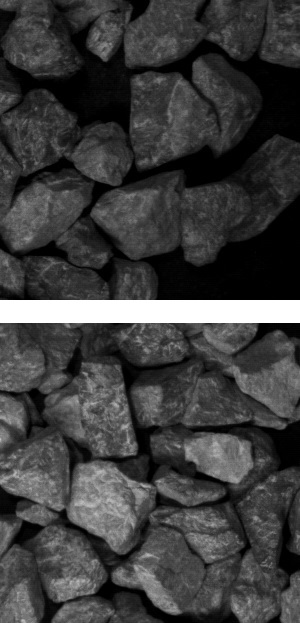}
    \label{ore}}
    \caption{Experiment equipment and acquired ore images. (a) Experimental platform. (b) Ore images with different densities. }
\end{figure}

\section{Experiments}\label{sec:experiments}
In this section, the ore dataset, evaluation metrics, and implementation details are first introduced. Then, ablation research is conducted to clarify the numerous design decisions. Finally, the proposed method is compared with state-of-the-art instance segmentation methods.

\subsection{Experiments Setup}

\subsubsection{Dataset}
For MS COCO~{\citep{COCO}} dataset, the networks are trained with train2017 (115K images) and evaluated with val2017 (5K images).
In industrial production, detecting the particle size of some ores can determine the particle size distribution of a batch of ores. So, two-dimensional image detection can achieve practical application results. The surface ore on the conveyor belt is used as a test sample, and its accuracy must be ensured. Only independent and complete ore individuals are marked when creating a dataset. The ore image dataset is supplemented in the following ways: First, ore images are gathered at various scales with a resolution of 1280 $\times$ 1024 through the experiment platform, as shown in Fig. {\ref{platform}}. Then, the positioning of ores on various scales is altered, such as sparse, thick, etc. Furthermore, huge images are partitioned into a resolution of 300 $\times$ 300 images that can be used in network training. To boost the number of datasets even further, images are split with a sliding window in Fig. {\ref{ore}}. Finally, annotate these images in a standard COCO dataset format to evaluate network performance.
For the ore image dataset, 4060 images are used for training and 1060 for evaluation.
The ore image dataset is used to evaluate the accuracy of publicly available instance segmentation networks. For example, CondInst achieved the accuracy of 52.2 and 48.7 on the $ AP_{50}^{box} $ and $ AP_{50}^{mask} $, respectively. Its $ AP_{50}^{mask} $ on the COCO dataset reaches 56.4. This indicates that the experimental results obtained on the ore image dataset are reasonable.

\subsubsection{Evaluation Metrics}
To validate the effectiveness of the proposed method, 14 indexes were used:  $ AP^{box} $, $ AP_{50}^{box} $, $ AP_{75}^{box} $, $ AP^{mask} $, $ AP_{50}^{mask} $, $ AP_{75}^{mask} $, $ AR^{box} $, $ AR^{mask} $, Dice coefficient (Dice), training time, training memory consumption, inference memory usage, inference time (IT), and model size(MS). The top nine indicators are used to assess accuracy. 
$ AP^{box} $ and $ AP^{mask} $ represent the average accuracy of the detection box and mask, respectively, which can indicate the overall accuracy level of the network. $ AP_{50}^{box} $, $ AP_{50}^{mask} $,$ AP_{75}^{box} $, and $ AP_{75}^{mask} $ represent the detection box accuracy and mask accuracy when IoU is greater than 0.5 and 0.75, respectively. The larger the IoU, the more accurate the network is in locating the target. $ AR^{box} $ and $ AR^{mask} $ represent the average recall rates of the detection box and mask, respectively, which is the evaluation indicator for the missed detection of the model. In general, average precision (AP) and average recall (AR) are diametrically opposed, so it is critical to balance these two evaluation indicators.
The computational cost of the network is represented in the inferred time. The use of memory by a network during training illustrates the reliance of the network on the hardware system. The latter three indices are crucial in deciding if the model can be deployed on low-cost hardware. To ensure fairness in inference time, the inference time in the table is obtained by conducting multiple tests on 1000 images and calculating the mean of the results. To better compare the results, $ AR^{box} $ and $ AR^{mask} $ are counted at maxDets (max detections) = 10 in the ablation experiments and maxDets=100 in the comparison experiments.

\subsubsection{Implementation Details}
OreInst is trained with stochastic gradient descent (SGD) {\citep{8417976}}. For ore image dataset, all networks shown in Table {\ref{sota1}} are trained for 18k iterations on a single NVIDIA GTX3090 GPU with an initial learning rate of 0.01, which is then divided by 10 at 11.9k and again at 14.5k iterations. Image flipping and rotation are applied as strategies for data augmentation in network training. During training, the input images are modified to have shorter sides between [300, 480] pixels and longer sides between 640 pixels and less.
The network uses a random initialization method during training. And, the RGB mean and standard deviation of each image in the dataset should be calculated. The normalization method is that the corresponding average value should be subtracted from the RGB values of the input image and divided by the standard deviation.
No data augmentation is applied during testing, and the scales of the shorter and longer sides employed are 480 and 640, respectively. 
For MS COCO~\citep{COCO}, the networks are trained for 360K iterations on a single NVIDIA GTX3090 GPU with an initial learning rate of 0.005.
In the experiments in Section \ref{sec:ablation}, the batch size is 4. Especially on a single NVIDIA GTX2080Ti GPU, inference time and memory utilization are assessed.

\subsection{Ablation Study}\label{sec:ablation}
\subsubsection{Backbone}

To validate the effectiveness of the proposed framework, first, the overall structure of the network is examined by replacing the baseline network with various backbones. In Table \ref{backbone}, ResNet50 offers the best accuracy in $ AP_{50}^{mask} $, but its model size and inference time are 261 MB and 49.87 ms, respectively. 
Despite having the smallest model size, the ShuffleNetv2 is 6.15 less than MobileNetv3-small in $ AP_{50}^{box} $. And the ShuffleNetv2 is only 1.17 more than MobileNetv3-small in Dice. The $ AR^{box} $ and $ AR^{mask} $ are primarily used to explain the $ AP_{50}^{box} $ and $ AP_{50}^{mask} $ evaluation metrics rather than as the basis for selecting the backbone. Based on the comparison accuracy, model size, and inference time, MobileNetv3-small is the most suitable backbone.

\begin{table*}[!t]
	\renewcommand{\arraystretch}{1.25}
	\caption{Comparison of different backbones.
		\label{backbone}}
	\centering
		\small
		\setlength{\tabcolsep}{4.5mm}{
		\begin{tabular}{lccccccc}
			\toprule
			\multirow{2}{*}{\textbf{Backbone}}& 
			\multirow{2}{*}{\textbf{$AP_{50}^{box} $}}	& 
			\multirow{2}{*}{\textbf{$AR^{box} $}}	&
			\multirow{2}{*}{\textbf{$AP_{50}^{mask} $}}	& 
			\multirow{2}{*}{\textbf{$AR^{mask} $}}	&
			\multirow{2}{*}{\textbf{Dice}}	&
			\multirow{2}{*}{\begin{tabular}[c]{@{}c@{}}\textbf{IT}\\ \textbf{(ms)}\end{tabular}}&
			\multirow{2}{*}{\begin{tabular}[c]{@{}c@{}}\textbf{MS}\\ \textbf{(MB)}\end{tabular}}\\
			&&&&\\
			\midrule
			CSPDarkNet-53	&56.93 &41.9  &{\textbf{48.79}} &37.4 &42.34  &50.52  &283 \\
			DLA34	                        &57.52 &41.4  &48.58 &37.5 &42.33  &43.93  &193 \\
			EfficientNet-B0	&56.88 &41.2  &48.40 &37.4 &42.19  &48.27  &110 \\
			
			MobileNetv3-small &{\textbf{64.80}} &38.2  &47.93 &35.4 &40.72  &{\textbf{28.13}}  &90 \\
			ResNet50 	 &56.92 &{\textbf{42.2}}  &48.85 &{\textbf{37.9}} &{\textbf{42.68}}  &49.87  &261 \\
			ResNet101 &55.98 &{\textbf{42.2}}   &48.77 &{\textbf{37.9}}&42.65  &53.70  &406 \\
			PeleeNet	 &58.10 &40.3  &48.33 &36.9 &41.85  &49.10  &97 \\
			ShuffleNetv2         &58.65 &40.2  &48.28 &37.0 &41.89  &29.94  &{\textbf{84}} \\
			\bottomrule
	\end{tabular}}
\end{table*}

\begin{table*}[!t]
	\renewcommand{\arraystretch}{1.25}
	\caption{Ablation experiments associated with FPN.}
	\centering
	\label{FPN1}
	\setlength{\tabcolsep}{2pt}{
		\small
		\setlength{\tabcolsep}{4.5mm}{
		\begin{tabular}{ccccccccc}
			\toprule
			\multirow{2}{*}{\textbf{Channels}} & 
			\multirow{2}{*}{\textbf{Convs}} & 
			\multirow{2}{*}{\textbf{FEM}}  & 
			\multirow{2}{*}{$ AP_{50}^{box} $}      &  
			\multirow{2}{*}{$ AR^{box} $}      &  
			\multirow{2}{*}{$ AP_{50}^{mask} $}       & 
			\multirow{2}{*}{$ AR^{mask} $}      & 
			\multirow{2}{*}{\textbf{Dice}}      & 
			\multirow{2}{*}{\begin{tabular}[c]{@{}c@{}}\textbf{MS}\\ \textbf{(MB)}\end{tabular}}\\ 
			& & & & & &\\
			\midrule
			& \ding{51}    &\ding{55}      &66.95 &36.4  &47.35 &34.6 &39.98   &20.6 \\
			32      & \ding{55}    & \ding{55}     &{\textbf{66.96}}  &36.2  &47.31 &34.5 &39.90    &{\textbf{20.4}}\\
			&\ding{55}   	& \ding{51}    &65.74 &37.4  &47.64 &35.0 &40.35    &{\textbf{20.4}}\\
			\midrule
			& \ding{51}    &\ding{55}      &65.62 &37.6  &47.61 &35.0 &40.34  &25.0 \\
			64      & \ding{55}  	& \ding{55}    &65.75 &37.7  &47.87 &35.2 &40.57    &24.2\\
			&\ding{55}   	& \ding{51}    &64.40 &38.3  &48.02 &35.4 &40.76    &24.3\\
			\midrule
			& \ding{51}    &\ding{55}      &65.18 &38.1  &47.94 &35.3 &40.66  &31.3 \\
			96      & \ding{55}  	& \ding{55}    &64.89 &38.1  &48.00 &35.3 &40.68    &29.4\\
			&\ding{55}   	& \ding{51}    &63.98 &38.8  &48.12 &35.6 &40.98    &29.7\\
			\midrule
			& \ding{51}     &\ding{55}      &65.19 &38.1  &47.83 &35.3 &40.62  &39.4 \\
			128     & \ding{55}  	& \ding{55}    &65.29 &38.3  &48.03 &35.4 &40.76    &36.0\\
			&\ding{55}   	& \ding{51}    &63.31 &38.9  &48.14 &35.6 &40.93    &36.5\\
			\midrule 
			& \ding{51}    &\ding{55}       &64.33 &38.3  &47.93 &35.3 &40.66  &49.4 \\
			160     & \ding{55}    & \ding{55}    &64.83 &38.4  &47.97 &35.5 &40.80    &44.1\\
			&\ding{55}     & \ding{51}    &63.72 &39.1  &48.21 &{\textbf{35.7}} &{\textbf{41.02}}   &44.7\\
			\midrule 
			& \ding{51}    &\ding{55}      &64.80 &38.2  &47.93 &35.4 &40.72  &90.1 \\
			256     & \ding{55}  	& \ding{55}    &64.64 &38.4  &48.07 &35.4 &40.77    &76.6\\
			&\ding{55}   	& \ding{51}    &63.41 &{\textbf{39.1}}  &{\textbf{48.21}}&35.6 &40.96   &78.2\\
			\bottomrule
	\end{tabular}}}
\end{table*}

\subsubsection{Ghost-FPN}
The number of input channels in FPN is reduced to compress the model size. In Table \ref{FPN1}, When the number of channels is reduced from 256 and 96 to 96 and 32, the model size is reduced to 33$\%$ and 66$\%$, respectively. But when the number of channels is 96 and 32, respectively, the latter is 0.59 lower than the former in $ AP_{50}^{mask} $.
The above data shows that channel 96 is the most optimal. On this basis, the model size and accuracy are compressed and maintained by removing the output convolutions and adding FEM. After these two operations, the accuracy, Dice, and model size have been improved.
Due to the characteristics of ore images, most of the small-scale information in FPN is redundant. In Table {\ref{FPN2}}, eliminating P6 and P7 reduces the model size and Dice from 29.7 MB and 40.92 to 28.4 MB and 41.00, respectively.


\begin{table*}[!t] 
	\renewcommand{\arraystretch}{1.25}
	\caption{Ablation study of P6$\&$P7 layer in FPN.}
	\centering
	\label{FPN2}
	\small
	\setlength{\tabcolsep}{5mm}{
		\begin{tabular}{ccccccc}
		\toprule
		\multirow{2}{*}{\textbf{P6 $\&$ P7}}  & 
		\multirow{2}{*}{$ AP_{50}^{box} $}      &  
		\multirow{2}{*}{$ AR^{box} $}       & 
		\multirow{2}{*}{$ AP_{50}^{mask} $}       & 
		\multirow{2}{*}{$ AR^{mask} $}       & 
		\multirow{2}{*}{\textbf{Dice}}       & 
		\multirow{2}{*}{\begin{tabular}[c]{@{}c@{}}\textbf{MS}\\ \textbf{(MB)}\end{tabular}}\\ 
		& & & & \\
		\midrule
		\ding{51}		&63.98 &38.8  &48.12 &35.6 &40.92    &29.7   \\
		\ding{55}		&63.65 &38.9  &48.16 &35.7 &41.00    &28.4   \\
		\bottomrule
	\end{tabular}}
\end{table*}

\subsubsection{Optimized of Detection Head}
Table {\ref{FCOS1}} demonstrates that as the number of convolutional layers is reduced, mask accuracy and model size decrease. When only the number of convolution layers is changed and is reduced from 4 to 2, $ AP_{50}^{mask} $, Dice, and model size are reduced by 0.64, 0.82, and 2.5 MB, and $ AP_{50}^{box} $ is improved by 1.92. When the number of convolution layers is reduced from 2 to 1, $ AP_{50}^{mask} $, Dice, and model size are reduced by 0.73, 0.73, and 1.3 MB, and $ AP_{50}^{box} $ is improved by 1.07. When the convolution layer remains unchanged, model size is reduced even further after employing shared convolution (SC) and depthwise separable convolution (DSC), while the accuracy remains nearly unchanged.
According to the four evaluation indexes in the table, the suitable option is two convolution layers and uses shared convolution and depthwise separable convolution.


\begin{table*}[!t]
	\renewcommand{\arraystretch}{1.25}
	\caption{Ablation experiments associated with detection heads. Layer means the number of convolutional layers. }
	\centering
	\label{FCOS1}
	\small
	\setlength{\tabcolsep}{5mm}{
	\begin{tabular}{cccccccccc}
		\toprule
		\multirow{2}{*}{\textbf{Layer}} & 
		\multirow{2}{*}{\textbf{SC}} & 
		\multirow{2}{*}{\textbf{DSC}}  & 
		\multirow{2}{*}{$ AP_{50}^{box} $}      &  
		\multirow{2}{*}{$ AR^{box} $}      &  
		\multirow{2}{*}{$ AP_{50}^{mask} $}       & 
		\multirow{2}{*}{$ AR^{mask} $}      & 
		\multirow{2}{*}{\textbf{Dice}}      & 
		\multirow{2}{*}{\begin{tabular}[c]{@{}c@{}}\textbf{MS}\\ \textbf{(MB)}\end{tabular}}\\ 
		& & & & & & \\
		\midrule
		& \ding{55}     & \ding{55}&66.64 &34.6  &46.79 &34.1 &39.45  &24.6 \\
		1      	& \ding{51}  	& \ding{55}&{\textbf{66.66}} &34.4  &46.68 &34.0 &39.34   &24.0\\
		& \ding{51}  	& \ding{51}     &66.45 &35.2   &46.90 &34.3 &39.62    &{\textbf{23.4}}\\
		\midrule
		& \ding{55}    & \ding{55} &65.57 &36.6  &47.52 &34.8 &40.18    &25.9\\
		2       & \ding{51}  	&\ding{55} &65.40 &36.3  &47.32 &34.6 &39.97    &24.6\\
		& \ding{51}  	& \ding{51}  &66.51 &36.6 &47.40 &34.5 &39.93    &23.5\\
		\midrule
		&\ding{55}     	&\ding{55}   &64.18 &38.0   &48.04 &35.3 &40.70    &27.1\\
		3       & \ding{51}  	&\ding{55}   &64.53 &37.9   &47.77 &35.3 &40.60    &25.2\\
		& \ding{51}  	& \ding{51}  &65.82 &37.6 &47.53 &35.2 &40.45    &23.6\\
		\midrule
		&\ding{55}      &\ding{55}   &63.65 &{\textbf{38.9}}   &48.16 &{\textbf{35.7}} &41.00    &28.4\\
		4       & \ding{51}  	&\ding{55}    &64.34 &38.8    &{\textbf{48.19}}&{\textbf{35.7}} &{\textbf{41.02}}   &25.9\\
		& \ding{51}  	& \ding{51} &65.79 &38.4  &48.02 &35.5 &40.82    &23.7\\ 
		\bottomrule 
	\end{tabular}}
\end{table*}

\subsubsection{Fusion Feature Similarity}
Table {\ref{colorspace}} contrasts the performance of several color spaces and other features. In Table {\ref{colorspace}}, 
The Lab outperforms the other color spaces in $ AP_{50}^{mask} $ and Dice. When compared to the LBP texture feature and the histogram of oriented gradients (HOG) feature, the former performs better than the latter. But the Lab is still combined with the LBP texture feature or HOG feature at varied weights for further ablation experiments to demonstrate the validity of the choice. The best performance is obtained when the weights of the Lab and LBP are 0.9 and 0.1, respectively. The change from pixel to similarity computation foundation has no effect on inference time or $ AP_{50}^{box} $, the $ AP_{50}^{mask} $, and Dice is improved without any loss.

In Table {\ref{coco}}, the proposed fusion feature outperforms the Lab in $ AP^{mask} $, $ AP_{50}^{mask} $, $ AP_{75}^{mask} $, $ AP^{box} $ and $ AP_{75}^{box} $ on the MS COCO dataset . This demonstrates that the proposed fusion features have more potential for discussion on the coco dataset.

\begin{table*}[!t]
	\renewcommand{\arraystretch}{1.25}
	\caption{Comparison of different color spaces and other features.
		\label{colorspace}}
	\centering
	\small 
	\setlength{\tabcolsep}{5mm}{
		\begin{tabular}{ccccccc}
		\toprule
		\textbf{Feature} & \textbf{Weight}  & $ AP_{50}^{box} $  & $ AR^{box} $     &  $ AP_{50}^{mask} $ & $ AR^{mask} $  & \textbf{Dice}\\
		\midrule
		Lab	&- &66.51 &36.6 &47.40 &34.5  &39.93 \\
		XYZ		&- &66.49 &30.6 &43.22 &23.6 &30.53 \\
		YUV			&- &66.66 &31.9 &45.07 &26.4 &33.30 \\
		HSV		&- &{\textbf{68.30}} &31.9 &45.23 &27.6 &34.28 \\
		LBP 		&- &66.50 &32.6 &45.62 &25.9 &33.04 \\
		HOG		&- &67.55 &31.5 &44.38 &22.2 &29.60 \\
		\midrule
		&0.7 $\&$ 0.3 &66.73 &{\textbf{37.0}} &47.21 &32.3 &38.36 \\
		Lab $\&$ LBP 		&0.8 $\&$ 0.2 &66.44 &36.6 &47.39  &34.5 &39.93\\
		&0.9 $\&$ 0.1 &66.93 &36.7 &{\textbf{47.67}}&{\textbf{34.6}} &{\textbf{40.10}} \\
		&0.7 $\&$ 0.3 &67.85 &33.8 &46.14 &24.1 &31.66 \\
		Lab $\&$ HOG 		&0.8 $\&$ 0.2 &67.31 &33.8 &46.84 &30.6 &37.02 \\
		&0.9 $\&$ 0.1 &66.42 &36.3 &47.45 &34.1 &39.68 \\
		\bottomrule
	\end{tabular}}
\end{table*}

\begin{table*}[!t]
	\renewcommand{\arraystretch}{1.25}
 	\setlength{\tabcolsep}{5mm}{
	\caption{Comparison Lab with Lab $\&$ LBP on MS COCO dataset.
		\label{coco}}
	\centering
	\small 
	\begin{tabular}{cccccccccc}
		\toprule
		\textbf{Feature} & \textbf{$ AP^{box} $}  & \textbf{$ AP_{50}^{box} $}      & \textbf{$ AP_{75}^{box} $} & \textbf{$ AP^{mask} $}  & \textbf{$ AP_{50}^{mask} $}      & \textbf{$ AP_{75}^{mask} $} \\
		\midrule
		Lab					&40.42 &59.03 &43.72 &31.15 &53.21 &31.65  \\
		Lab $\&$ LBP		&40.62 &58.95 &44.11 &31.24 &53.26 &31.85  \\
		\bottomrule
	\end{tabular}}
\end{table*}

\begin{table*}[!t]
	\renewcommand{\arraystretch}{1.25}
	\caption{Comparison with the state-of-the-art methods on ore image dataset.
		\label{sota1}}
	\centering
	\small
	\setlength{\tabcolsep}{0.7mm}{
		\begin{tabular}{lccccccccccc}
			\toprule
			\multirow{2}{*}{\textbf{Methods}}& 
			\multirow{2}{*}{\textbf{Backbone}}& 
			\multirow{2}{*}{\textbf{$AP_{50}^{box}\uparrow$}}& 
			\multirow{2}{*}{\textbf{$AR^{box}\uparrow$}}	&
			\multirow{2}{*}{\textbf{$AP_{50}^{mask}\uparrow $}}	& 
			\multirow{2}{*}{\textbf{$AR^{mask}\uparrow$}}	&
			\multirow{2}{*}{\textbf{Dice$\uparrow$}}	&
			\multicolumn{2}{c}{\textbf{Time(ms)}} & 
			\multicolumn{2}{c}{\textbf{Memory(MB)}} &
			\multirow{2}{*}{\begin{tabular}[c]{@{}c@{}}\textbf{Model}\\ \textbf{size(MB)}\end{tabular}}\\
			& & & & & & &Train &Infer &Train &Infer & \\
			\midrule
			\emph{Fully-supervised}: & & & & & & & & & & &\\
			Mask R-CNN 	&ResNet101 &51.7 &45.9 &22.1 &24.7 &23.3 &251.2 &74.4 &5407 &1876 &480\\
			Mask R-CNN  &ResNet50  &51.7 &45.7 &22.6 &24.9 &23.7 &361.7 &60.1 &6499 &1769 &334\\ 
			YOLACT      &ResNet101 &50.6 &44.1 &11.9 &18.6 &14.5 &82.7 &40.9 &10001 &7758 &410\\
			YOLACT      &ResNet50  &50.5 &43.6 &11.5 &18.0 &14.0 &{\textbf{64.0}} &32.8 &10513 &7610 &265\\
			HTC         &ResNet50  &52.6 &46.9 &49.5 &43.0 &{\textbf{46.0}}&893.4 &78.9 &6165 &2314 &588\\
			SOLOv2      &ResNet50  &-    &- &45.8 &36.7 &40.8 &392.7 &45.3 &5025 &3123 &354\\
			CondInst    &ResNet50  &52.2 &47.2 &48.7 &{\textbf{43.3}} &45.8 &312.8 &42.1 &5385 &1713 &259\\
			CARAFE      &ResNet50  &51.8 &45.8 &24.9 &25.8 &25.3 &515.7 &65.2 &7373 &1873 &376\\
			MS R-CNN    &ResNet50 &51.7 &45.7 &23.4 &24.9 &24.1 &246.6 &62.7 &7249 &1749 &428\\
			BlendMask   &ResNet50 &58.1 &50.0 &48.7 &43.1 &45.7 &276.9 &43.4 &4669 &1641 &274\\
			\midrule
			\emph{Box-supervised}: & & & & & & & & & & &\\
			BBTP     	&ResNet50 &51.6 &42.8 &48.0 &29.9 &36.9 &690.0 &45.6 &{\textbf{3643}} &1514 &334\\
			BBTP  		&ResNet101 &51.7 &45.2 &48.3 &26.7 &34.4 &765.1 &66.8 &5234 &1820 &479\\
	    	BoxInst     &ResNet50 &56.9 &49.2 &48.8 &41.9 &45.1 &454.6 &42.7 &9276 &1681 &261\\
			DiscoBox  	&ResNet50&{\textbf{73.8}} &{\textbf{82.5}} &49.0 &42.2 &45.4 &743.0 &88.7 &9093 &2405 &352\\
			OreInst (ours) 	         	&ResNet50 &62.0 &50.3 &{\textbf{49.3}} &41.6 &45.1 &841.3 &24.8 &8430 &1549 &190 \\
			OreInst (ours) 		        &MobileNetv3-small &67.8 &53.6 &47.7 &39.9 &43.5 &639.4 &{\textbf{19.8}}&7585 &{\textbf{1431}} &{\textbf{23.5}}\\
			\bottomrule
	\end{tabular}}
\end{table*}

\begin{figure*}[!t]
	\centering
	\subfigure[Input]{
		\begin{minipage}[b]{0.15\linewidth} 
		\includegraphics[width=0.9in]{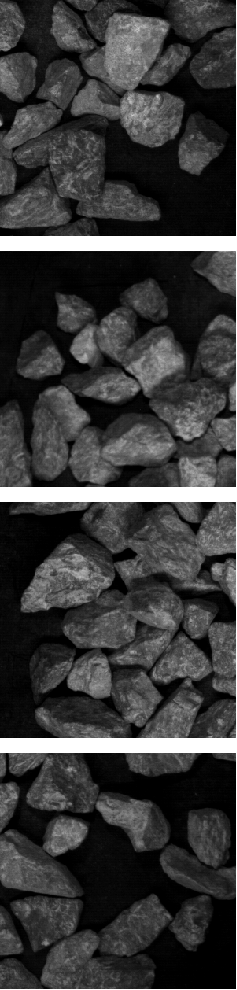}\vspace{1pt} 
		\end{minipage}
	}
	\subfigure[Ground truth]{
		\begin{minipage}[b]{0.15\linewidth}
		\includegraphics[width=0.9in]{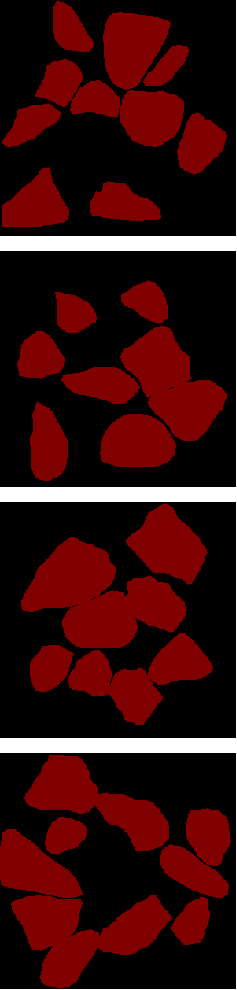}\vspace{1pt} 

		\end{minipage}
	}
	\subfigure[Mask R-CNN]{
		\begin{minipage}[b]{0.15\linewidth} 
		\includegraphics[width=0.9in]{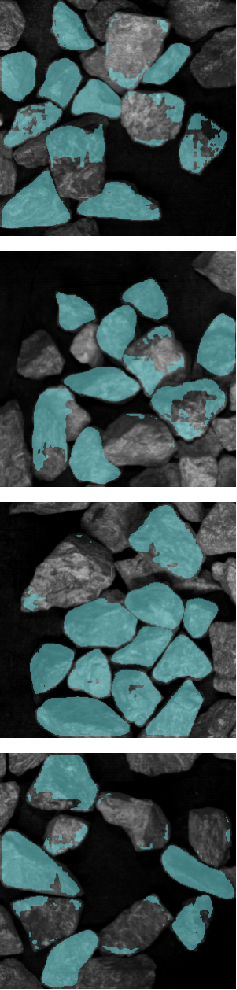}\vspace{1pt} 

		\end{minipage}
	}
	\subfigure[YOLACT]{
		\begin{minipage}[b]{0.15\linewidth}
		\includegraphics[width=0.9in]{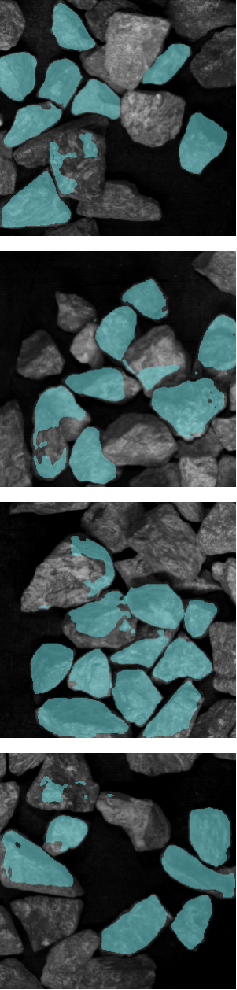}\vspace{1pt} 
 
		\end{minipage}
	}

	\centering
	\subfigure[DiscoBox]{
		\begin{minipage}[b]{0.15\linewidth}
			\includegraphics[width=0.9in]{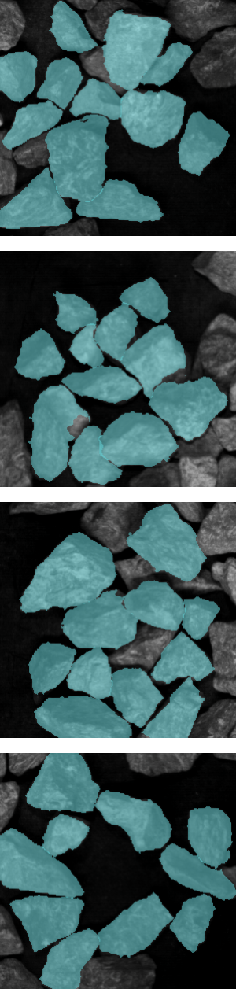}\vspace{1pt} 
			 
		\end{minipage}
	}
	\subfigure[BoxInst]{
		\begin{minipage}[b]{0.15\linewidth}
			\includegraphics[width=0.9in]{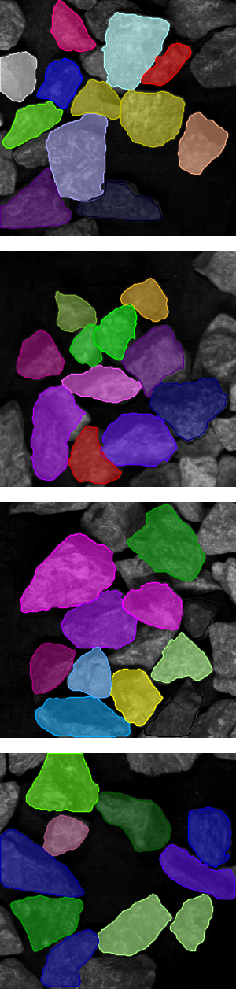}\vspace{1pt} 
	 
		\end{minipage}
	}
	\subfigure[OreInst-R50]{
		\begin{minipage}[b]{0.15\linewidth}
			\includegraphics[width=0.9in]{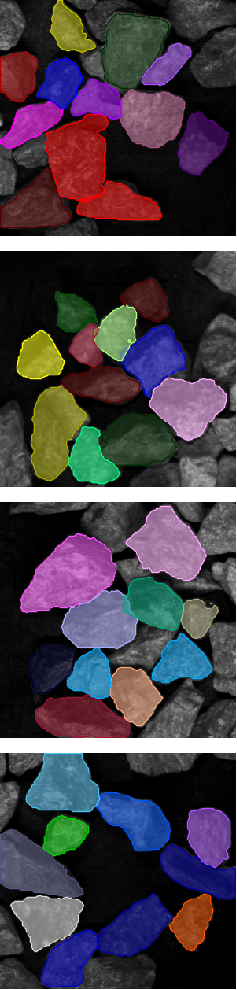}\vspace{1pt} 
 
		\end{minipage}
	}
	\subfigure[OreInst-MV3]{
		\begin{minipage}[b]{0.15\linewidth}
			\includegraphics[width=0.9in]{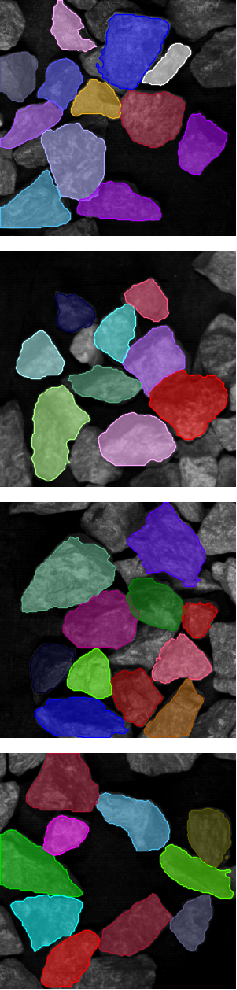}\vspace{1pt} 

		\end{minipage}
	}
	\caption{The visual results of ore images processed by different methods.}
	\label{pics}
\end{figure*}

\subsection{Comparison with State-of-the-Art Methods}
To further illustrate the superiority of OreInst, the proposed framework is compared with state-of-the-art methods. Using the ResNet50, $AP_{50}^{mask} $ and Dice of OreInst reach 49.3 and 45.1, which is only 0.2 and 0.9 lower than that of HTC. The application of FEM, depthwise separable convolution, and the proposed fusion feature contribute to the high accuracy of OreInst. Similarly, taking the MobileNetv3-small, the inference time and model size of OreInst are far smaller than those of any other methods, which is due to the lightweight backbone MobileNetv3-small and Ghost-FPN, as well as the optimized detection head. In the meantime, OreInst maintain 67.8 in $AP_{50}^{box} $ , 47.7 in $AP_{50}^{mask} $, and 43.5 in Dice with MobileNetv3-small.

In the fully-supervised approaches, HTC and YOLACT have the best $AP_{50}^{mask} $ and training time. However, the $AP_{50}^{box} $, $AP_{50}^{mask} $, Dice of YOLACT are 10.3, 37.7, and 31.1 lower than OreInst with the same backbone ResNet50, respectively. And the inference time and model size of OreInst are 8 ms and 67 MB lower than that of YOLACT, respectively. In the same case, both inference time and model size of OreInst are only 31$\%$ and 32$\%$ of those of HTC, respectively. If MobileNetv3-small is used, OreInst significantly decreases inference time and model size. In the box-supervised approaches, DiscoBox has the best $AP_{50}^{box} $ and Dice, which performs significantly worse than OreInst with ResNet50 in the critical indicators of $AP_{50}^{mask} $, inference time and model size. With MobileNetv3-small as the backbone, the inference time and model size of OreInst are only 22$\%$ and 6$\%$ of those of DiscoBox, respectively. And for BBTP, only training memory is better than OreInst, while all other indicators are worse. According to the comparison above, OreInst performs better. Figure {\ref{pics}} shows that OreInst performs admirably regarding segmentation impact, which is superior for border segmentation of the lowest two ores in the second picture.

\subsection{Limitation and Discussion}\label{sec:discussion}
\subsubsection{Limitation}
The work in the article still has certain limitations. These limitations are broken down into restrictions during network use and restrictions during deployment. Regarding the limitations in the use of the network, although images are subjected to data augmentation operations such as scaling, rotation, and flipping before being input for network training.  The uniform pixel size of the input image, on the other hand, restricts the capacity of the network to detect images with pixel sizes that differ from the training image. Similarly, the accuracy of network will be limited if the lighting conditions of the detected image differ from those in the trained dataset. These limitations can be overcome by gathering images with various pixel sizes and altering the lighting situation.
In terms of deployment, the paradigm still has several limits and downsides. When there is stacking between ores, the missed detection rate of the network increases in practice. The camera can only get complete and independent individuals from the upper layer because the detection object is a flat image. This problem can be alleviated by using a scientific mechanical structure in the system to treat the ore stacked on the conveyor belt. 3D image algorithms can tackle the problem of stacking between targets at the algorithmic level. Second, the network struggles to find their edges reliably when ores with nearly identical color features cling together. Based on the detection box, the network can distinguish sticky objects from various individuals, but edges are difficult to identify reliably. This difficulty can be mitigated by using scientific light source settings. Third, the annotation process of the dataset still requires some time. This challenge can be effectively solved using self-supervised or unsupervised training techniques. Finally, while the inference process takes very little memory, memory utilization during training is significant.
\subsubsection{Discussion}
Images without ores were excluded from the dataset when it was created. However, in practical work, the images captured by the camera may not contain minerals. In this case, the instability of image data leads to unstable accuracy of the network. And there is the instability of inference time. When calculating inference time, there are certain fluctuations in the inference time of a single image. To ensure fairness in inference time, the inference time in the table is obtained by conducting multiple tests on 1000 images and calculating the mean of the results. However, in the actual detection process, the network cannot always maintain a stable inference speed when processing each image. The unstable inference speed may be caused by both the instability of the hardware and the number of targets in the image. The inference speed will be more unstable when dealing with continuous real-time data feeding. The instability of hardware and image datasets leads to uncertainty in inference results. This kind of uncertainty persists during the whole model training and model application process. This uncertainty needs to be analyzed and discussed \citep{uncerten1,uncerten2}, and ways to prevent it need to be discovered.
Image buffering technology is frequently used in industrial settings to address this issue, which has specific performance requirements for cameras. The issue of inconsistent speed can also be resolved by batch-processing images, but results feedback may be delayed as a result. Meanwhile, differences between the gathered and training images may result from environmental changes. Applying a lightweight network to numerous scenarios with notable differences is challenging. Samples should be manually extracted to check whether the network is still appropriate for the current situation when the data and environment have changed. The ability of networks to generalize is uncertain in this sense.

When annotating the ore dataset, obscured ores will not be annotated. In Fig. {\ref{pics}} of the article, the ground truth only includes the complete ores. Based on this annotation principle, the features of the obscured ores will not be learned during the network training process. The obscured ores are excluded from the target of network learning. So, the input image needs to have the following constraints. Firstly, the input image size should be close to the image size used for training, and secondly, the input image should be clearly focused. The input image with the above constraints can achieve the most ideal results.

Some new technologies should be taken into account when more complex scenarios make segmentation tasks challenging. Multitask deep learning can fulfill the requirements when multiple tasks must be completed simultaneously in the real world. 
In a panoptic driving perception network, you only look once for panoptic (YOLOP)~{\citep{wu2022yolop}} to perform traffic object detection, drivable area segmentation, and lane detection simultaneously. This method is well suited for this particular application scenario when the ore is moving on the conveyor belt, and the input is in video format.
A proposed adaptive multi-scale feature fusion-based multi-task semantic segmentation network architecture~{\citep{Multitask1}} enhances small-scale target segmentation accuracy and segmentation target edge details by combining boundary detection and semantic segmentation tasks. With this approach, the issue of small target ores and stringent edge segmentation accuracy requirements can be resolved. To optimize network performance when parameter adjustment becomes complex, which is not a simple task for regular people to do, professionals are required. Recent research has used automated deep learning techniques to successfully detect and segment medical images~{\citep{auto1}}. It integrated several state-of-the-art CNN-based architectures in order to obtain a fully automated pipeline for the complete segmentation and classification via Deep Learning techniques, in which the results of segmentation are used to improve classification results and vice versa in a mutual and cyclic way. In scenarios involving industrial applications, these techniques merit further investigation.

\section{Conclusion}\label{sec:conclusion}
In this paper, an efficient segmentation with texture in ore images named OreInst based on a box-supervised approach is developed. The proposed OreInst includes a lightweight structural design and a loss function based on fusion features, allowing the proposed framework to be lightweight and run on hardware with limited resources while maintaining accuracy. The experiments on the MS COCO dataset show that the proposed fusion features with texture can improve the performance of instance segmentation. Experiments on ore image dataset demonstrate that OreInst can achieve a real-time speed of over 50 FPS with a small model size of 23.5 MB. Meanwhile, OreInst retains competitive accuracy performance, i.e., 67.8 in $ AP_{50}^{box} $ and 47.7 in $ AP_{50}^{mask} $, in comparison with the state-of-the-art methods. The proposed method has the advantages of a light model with fast detection speed and can be effectively applied to mineral processing operations. For future work, we will focus on accurately obtaining ore edge information to improve accuracy and further expand the proposed approach to other industrial fields. At the same time, the design of the mineral processing system is explored to solve the problem of real-time data generation.
\\

\bibliographystyle{plainnat}

\bibliography{cas-refs}

\begin{thebibliography}{40}
\providecommand{\natexlab}[1]{#1}
\providecommand{\url}[1]{\texttt{#1}}
\expandafter\ifx\csname urlstyle\endcsname\relax
  \providecommand{\doi}[1]{doi: #1}\else
  \providecommand{\doi}{doi: \begingroup \urlstyle{rm}\Url}\fi

\bibitem[Al-Huda et~al.(2023)Al-Huda, Peng, Algburi, Al-antari, AL-Jarazi, and
  Zhai]{EAAI5}
Zaid Al-Huda, Bo~Peng, Riyadh Nazar~Ali Algburi, Mugahed~A. Al-antari, Rabea
  AL-Jarazi, and Donghai Zhai.
\newblock A hybrid deep learning pavement crack semantic segmentation.
\newblock \emph{Engineering Applications of Artificial Intelligence},
  122:\penalty0 106142, 2023.
\newblock ISSN 0952-1976.

\bibitem[Amankwah and Aldrich(2011)]{2011Automatic}
Anthony Amankwah and Chris Aldrich.
\newblock Automatic ore image segmentation using mean shift and watershed
  transform.
\newblock In \emph{Proceedings of 21st International Conference
  Radioelektronika}, pages 245 -- 248, 2011.

\bibitem[Asheghi et~al.(2020)Asheghi, Hosseini, Saneie, and Shahri]{ann}
Reza Asheghi, Seyed~Abbas Hosseini, Mojtaba Saneie, and Abbas~Abbaszadeh
  Shahri.
\newblock Updating the neural network sediment load models using different
  sensitivity analysis methods: a regional application.
\newblock \emph{Journal of Hydroinformatics}, 22:\penalty0 562--577, 2020.

\bibitem[Bolya et~al.(2019)Bolya, Zhou, Xiao, and Lee]{YOLACT}
Daniel Bolya, Chong Zhou, Fanyi Xiao, and Yong~Jae Lee.
\newblock {YOLACT:} real-time instance segmentation.
\newblock In \emph{IEEE International Conference on Computer Vision}, pages
  9156--9165, 2019.

\bibitem[Chalfoun et~al.(2014)Chalfoun, Majurski, Dima, Stuelten, Peskin, and
  Brady]{2014FogBank}
Joe Chalfoun, Michael Majurski, Alden Dima, Christina Stuelten, Adele Peskin,
  and Mary Brady.
\newblock Fogbank: A single cell segmentation across multiple cell lines and
  image modalities.
\newblock \emph{Bioinformatics}, 15\penalty0 (1):\penalty0 431, 2014.

\bibitem[Chen et~al.(2020)Chen, Sun, Tian, Shen, Huang, and Yan]{BlendMask}
Hao Chen, Kunyang Sun, Zhi Tian, Chunhua Shen, Yongming Huang, and Youliang
  Yan.
\newblock Blendmask: Top-down meets bottom-up for instance segmentation.
\newblock In \emph{IEEE Conference on Computer Vision and Pattern Recognition},
  pages 8570--8578, 2020.

\bibitem[Chen et~al.(2022{\natexlab{a}})Chen, Yang, and Lyu]{Multitask1}
Huilin Chen, Shengsong Yang, and Ting Lyu.
\newblock Multitask semantic segmentation network using adaptive multiscale
  feature fusion.
\newblock pages 64--69, 2022{\natexlab{a}}.

\bibitem[Chen et~al.(2019)Chen, Pang, Wang, Xiong, Li, Sun, Feng, Liu, Shi,
  Ouyang, Loy, and Lin]{HTC}
Kai Chen, Jiangmiao Pang, Jiaqi Wang, Yu~Xiong, Xiaoxiao Li, Shuyang Sun,
  Wansen Feng, Ziwei Liu, Jianping Shi, Wanli Ouyang, Chen~Change Loy, and
  Dahua Lin.
\newblock Hybrid task cascade for instance segmentation.
\newblock In \emph{IEEE Conference on Computer Vision and Pattern Recognition},
  pages 4969--4978, 2019.

\bibitem[Chen et~al.(2022{\natexlab{b}})Chen, Zhao, and Zhang]{eaai1}
Long Chen, Yin-Ping Zhao, and Chuanbin Zhang.
\newblock Efficient kernel fuzzy clustering via random fourier superpixel and
  graph prior for color image segmentation.
\newblock \emph{Engineering Applications of Artificial Intelligence},
  116:\penalty0 105335, 2022{\natexlab{b}}.
\newblock ISSN 0952-1976.

\bibitem[Han et~al.(2020)Han, Wang, Tian, Guo, Xu, and Xu]{GhostNet}
Kai Han, Yunhe Wang, Qi~Tian, Jianyuan Guo, Chunjing Xu, and Chang Xu.
\newblock Ghostnet: More features from cheap operations.
\newblock In \emph{IEEE Conference on Computer Vision and Pattern Recognition},
  pages 1577--1586, 2020.

\bibitem[He et~al.(2017)He, Gkioxari, Dollar, and Girshick]{Mask-RCNN}
Kaiming He, Georgia Gkioxari, Piotr Dollar, and Ross Girshick.
\newblock {Mask R-CNN}.
\newblock In \emph{IEEE International Conference on Computer Vision}, pages
  2980--2988, 2017.

\bibitem[Hosseini et~al.(2022)Hosseini, Shahri, and Asheghi]{Block_network1}
Seyed~Abbas Hosseini, Abbas~Abbaszadeh Shahri, and Reza Asheghi.
\newblock Prediction of bedload transport rate using a block combined network
  structure.
\newblock \emph{Hydrological Sciences Journal}, 67\penalty0 (1):\penalty0
  117--128, 2022.

\bibitem[Howard et~al.(2019)Howard, Sandler, Chen, Wang, Chen, Tan, Chu,
  Vasudevan, Zhu, Pang, Adam, and Le]{MobileNetV3-Small}
Andrew Howard, Mark Sandler, Bo~Chen, Weijun Wang, Liang-Chieh Chen, Mingxing
  Tan, Grace Chu, Vijay Vasudevan, Yukun Zhu, Ruoming Pang, Hartwig Adam, and
  Quoc Le.
\newblock Searching for mobilenetv3.
\newblock In \emph{IEEE International Conference on Computer Vision}, pages
  1314--1324, 2019.

\bibitem[Hsu et~al.(2020)Hsu, Hsu, Tsai, Lin, Chuang, and Sinica]{BBTP}
Cheng-Chun Hsu, Kuang-Jui Hsu, Chung-Chi Tsai, Yen-Yu Lin, Yung-Yu Chuang, and
  Academia Sinica.
\newblock Weakly supervised instance segmentation using the bounding box
  tightness prior.
\newblock In \emph{Advances in Neural Information Processing Systems}, pages
  1--12, 2020.

\bibitem[Huang et~al.(2019)Huang, Huang, Gong, Huang, and
  Wang]{Mask-Scoring-RCNN}
Zhaojin Huang, Lichao Huang, Yongchao Gong, Chang Huang, and Xinggang Wang.
\newblock {Mask scoring R-CNN}.
\newblock In \emph{IEEE Conference on Computer Vision and Pattern Recognition},
  pages 6402 -- 6411, 2019.

\bibitem[Jamil and Roy(2023)]{pcgvit}
Sonain Jamil and Arunabha~M. Roy.
\newblock An efficient and robust phonocardiography (pcg)-based valvular heart
  diseases (vhd) detection framework using vision transformer (vit).
\newblock \emph{Computers in Biology and Medicine}, 158:\penalty0 106734, 2023.
\newblock ISSN 0010-4825.

\bibitem[Jiang et~al.(2022)Jiang, Chen, Wang, and Luo]{mglnn}
Bo~Jiang, Si~Chen, Beibei Wang, and Bin Luo.
\newblock Mglnn: Semi-supervised learning via multiple graph cooperative
  learning neural networks.
\newblock \emph{Neural Networks}, 153:\penalty0 204--214, 2022.
\newblock ISSN 0893-6080.

\bibitem[Khened et~al.(2021)Khened, Kori, Rajkumar, Krishnamurthi, and
  Srinivasan]{uncerten2}
Mahendra Khened, Avinash Kori, Haran Rajkumar, Ganapathy Krishnamurthi, and
  Balaji Srinivasan.
\newblock A generalized deep learning framework for whole-slide image
  segmentation and analysis.
\newblock \emph{Scientific Reports}, 11:\penalty0 11579, 2021.

\bibitem[Khoreva et~al.(2017)Khoreva, Benenson, Hosang, Hein, and Schiele]{SDI}
A.~Khoreva, R.~Benenson, J.~Hosang, M.~Hein, and B.~Schiele.
\newblock Simple does it: Weakly supervised instance and semantic segmentation.
\newblock In \emph{IEEE Conference on Computer Vision and Pattern Recognition},
  pages 1665--1674, 2017.

\bibitem[Krygier et~al.(2021)Krygier, LaBonte, Martinez, Norris, Sharma,
  Collins, Mukherjee, and Roberts]{uncerten1}
Michael~C. Krygier, Tyler LaBonte, Carianne Martinez, Chance Norris, Krish
  Sharma, Lincoln~N. Collins, Partha~P. Mukherjee, and Scott~A. Roberts.
\newblock Quantifying the unknown impact of segmentation uncertainty on
  image-based simulations.
\newblock \emph{Nature Communications}, 12:\penalty0 5414, 2021.

\bibitem[Lan et~al.(2021)Lan, Yu, Choy, Radhakrishnan, Liu, Zhu, Davis, and
  Anandkumar]{discobox}
Shiyi Lan, Zhiding Yu, Christopher Choy, Subhashree Radhakrishnan, Guilin Liu,
  Yuke Zhu, Larry~S. Davis, and Anima Anandkumar.
\newblock Discobox: Weakly supervised instance segmentation and semantic
  correspondence from box supervision.
\newblock In \emph{IEEE International Conference on Computer Vision}, pages
  3386--3396, 2021.

\bibitem[Lin et~al.(2014)Lin, Maire, Belongie, Hays, and Zitnick]{COCO}
T.~Y. Lin, M.~Maire, S.~Belongie, J.~Hays, and C.~L. Zitnick.
\newblock Microsoft coco: Common objects in context.
\newblock In \emph{European Conference on Computer Vision}, 2014.

\bibitem[Lin et~al.(2020)Lin, Goyal, Girshick, He, and Dollár]{8417976}
Tsung-Yi Lin, Priya Goyal, Ross Girshick, Kaiming He, and Piotr Dollár.
\newblock Focal loss for dense object detection.
\newblock \emph{IEEE Transactions on Pattern Analysis and Machine
  Intelligence}, 42\penalty0 (2):\penalty0 318--327, 2020.

\bibitem[Mukherjee et~al.(2009)Mukherjee, Potapovich, Levner, and
  Zhang]{2009Ore}
Dipti~Prasad Mukherjee, Yury Potapovich, Ilya Levner, and Hong Zhang.
\newblock Ore image segmentation by learning image and shape features.
\newblock \emph{Pattern Recognition Letters}, 30\penalty0 (6):\penalty0
  615--622, 2009.

\bibitem[Podda et~al.(2022)Podda, Balia, Barra, Carta, Fenu, and Piano]{auto1}
Alessandro~Sebastian Podda, Riccardo Balia, Silvio Barra, Salvatore Carta,
  Gianni Fenu, and Leonardo Piano.
\newblock Fully-automated deep learning pipeline for segmentation and
  classification of breast ultrasound images.
\newblock \emph{Journal of Computational Science}, 63:\penalty0 101816, 2022.

\bibitem[Rashedi and Nezamabadi-pour(2013)]{eaai3}
Esmat Rashedi and Hossein Nezamabadi-pour.
\newblock A stochastic gravitational approach to feature based color image
  segmentation.
\newblock \emph{Engineering Applications of Artificial Intelligence},
  26\penalty0 (4):\penalty0 1322--1332, 2013.
\newblock ISSN 0952-1976.

\bibitem[Ren et~al.(2017)Ren, He, Girshick, and Sun]{Faster-RCNN}
Shaoqing Ren, Kaiming He, Ross Girshick, and Jian Sun.
\newblock Faster r-cnn: Towards real-time object detection with region proposal
  networks.
\newblock \emph{IEEE Transactions on Pattern Analysis and Machine
  Intelligence}, 39\penalty0 (6):\penalty0 1137--1149, 2017.

\bibitem[Roy and Bhaduri(2023)]{densesph-yolov5}
Arunabha~M. Roy and Jayabrata Bhaduri.
\newblock Densesph-yolov5: An automated damage detection model based on
  densenet and swin-transformer prediction head-enabled yolov5 with attention
  mechanism.
\newblock \emph{Advanced Engineering Informatics}, 56:\penalty0 102007, 2023.
\newblock ISSN 1474-0346.

\bibitem[Shadmand and Mashoufi(2016)]{Block_network2}
Shirin Shadmand and Behbood Mashoufi.
\newblock A new personalized ecg signal classification algorithm using
  block-based neural network and particle swarm optimization.
\newblock \emph{Biomedical Signal Processing and Control}, 25:\penalty0 12--23,
  2016.

\bibitem[Tian et~al.(2020)Tian, Shen, and Chen]{CondInst}
Zhi Tian, Chunhua Shen, and Hao Chen.
\newblock Conditional convolutions for instance segmentation.
\newblock In \emph{European Conference on Computer Vision}, pages 282--298,
  2020.

\bibitem[Tian et~al.(2021)Tian, Shen, Wang, and Chen]{BoxInst}
Zhi Tian, Chunhua Shen, Xinlong Wang, and Hao Chen.
\newblock Boxinst: High-performance instance segmentation with box annotations.
\newblock In \emph{IEEE Conference on Computer Vision and Pattern Recognition},
  pages 5439--5448, 2021.

\bibitem[Wang et~al.(2019)Wang, Chen, Xu, Liu, Loy, and Lin]{CARAFE}
Jiaqi Wang, Kai Chen, Rui Xu, Ziwei Liu, Chen~Change Loy, and Dahua Lin.
\newblock {CARAFE:} content-aware reassembly of features.
\newblock In \emph{IEEE International Conference on Computer Vision}, pages
  3007--3016, 2019.

\bibitem[Wang et~al.(2023)Wang, Li, Zhang, and Fu]{EAAI6}
Wei Wang, Qing Li, Dezheng Zhang, and Jiawei Fu.
\newblock Image segmentation of adhesive ores based on msba-unet and
  convex-hull defect detection.
\newblock \emph{Engineering Applications of Artificial Intelligence},
  123:\penalty0 106185, 2023.
\newblock ISSN 0952-1976.

\bibitem[Wang et~al.(2020)Wang, Zhang, Kong, Li, and Shen]{SOLOv2}
Xinlong Wang, Rufeng Zhang, Tao Kong, Lei Li, and Chunhua Shen.
\newblock Solov2: Dynamic and fast instance segmentation.
\newblock In \emph{Advances in Neural Information Processing Systems}, pages
  1--17, 2020.

\bibitem[Wei et~al.(2023)Wei, Wei, Tang, Jia, Yin, and Ji]{EAAI4}
Dehua Wei, Xiukun Wei, Qingfeng Tang, Limin Jia, Xinqiang Yin, and Yang Ji.
\newblock Rtlseg: A novel multi-component inspection network for railway track
  line based on instance segmentation.
\newblock \emph{Engineering Applications of Artificial Intelligence},
  119:\penalty0 105822, 2023.
\newblock ISSN 0952-1976.

\bibitem[Wei et~al.(2021)Wei, Li, Xiao, Zhang, Miao, and Wang]{WangWei}
Wang Wei, Qing Li, Chengyong Xiao, Dezheng Zhang, Lei Miao, and Li~Wang.
\newblock An improved boundary-aware u-net for ore image semantic segmentation.
\newblock \emph{Sensors}, 21:\penalty0 2615, 2021.

\bibitem[Wu et~al.(2022)Wu, Liao, Zhang, Wang, Bai, Cheng, and
  Liu]{wu2022yolop}
Dong Wu, Man-Wen Liao, Wei-Tian Zhang, Xing-Gang Wang, Xiang Bai, Wen-Qing
  Cheng, and Wen-Yu Liu.
\newblock Yolop: You only look once for panoptic driving perception.
\newblock \emph{Machine Intelligence Research}, pages 1--13, 2022.

\bibitem[Yang et~al.(2012)Yang, Wang, Zhang, and Bu]{eaai2}
Hong-Ying Yang, Xiang-Yang Wang, Xian-Yin Zhang, and Juan Bu.
\newblock Color texture segmentation based on image pixel classification.
\newblock \emph{Engineering Applications of Artificial Intelligence},
  25\penalty0 (8):\penalty0 1656--1669, 2012.
\newblock ISSN 0952-1976.

\bibitem[Zhong et~al.(2018)Zhong, Yan, Wu, Shao, and Liu]{Block_network4}
Zhao Zhong, Junjie Yan, Wei Wu, Jing Shao, and Cheng-Lin Liu.
\newblock Practical block-wise neural network architecture generation.
\newblock In \emph{IEEE/CVF Conference on Computer Vision and Pattern
  Recognition}, pages 2423--2432, 2018.

\bibitem[Zhou et~al.(2017)Zhou, Ni, and Rao]{Block_network3}
Jianghong Zhou, Jiangqun Ni, and Yuan Rao.
\newblock Block-based convolutional neural network for image forgery detection.
\newblock In \emph{International Workshop on Digital Watermarking}, pages
  65--76, 2017.

\end{thebibliography}

\end{document}